\documentclass{article}

\PassOptionsToPackage{numbers, compress}{natbib}

\usepackage[main,final]{neurips_2026}

\usepackage{graphicx}
\usepackage{adjustbox}      
\usepackage{booktabs}
\usepackage{kotex}
\usepackage{natbib}

\usepackage{multirow}
\usepackage{xcolor}
\newcommand{\inc}[1]{\textcolor{blue}{#1}} 
\newcommand{\dec}[1]{\textcolor{red}{#1}}  
\newcommand{\eqv}[1]{\textcolor{gray}{#1}} 

\usepackage[accsupp]{axessibility}  

\usepackage[utf8]{inputenc} 
\usepackage[T1]{fontenc}    
\usepackage{hyperref}       
\usepackage{url}            
\usepackage{booktabs}       
\usepackage{amsfonts}       
\usepackage{nicefrac}       
\usepackage{microtype}      
\usepackage{xcolor}         

\usepackage{hyperref}

\usepackage[nameinlink,noabbrev,capitalize]{cleveref}
\crefname{figure}{Fig.}{Figs.}
\crefname{table}{Table}{Tables}
\crefname{section}{Sec.}{Secs.}
\crefname{equation}{Eq.}{Eqs.}

\newcommand{\etal}{\textit{et al.}}

\title{Grounding Driving VLA via Inverse Kinematics}

\author{%
  Junsung Park, Hyunjung Shim \\
  Korea Advanced Institute of Science and Technology\\
  Kim Jaechul Graduate School of AI\\
  18, Taebong-ro, Seocho-gu, Seoul, South Korea \\
  \texttt{\{jshackist, kateshim\}@kaist.ac.kr} \\
}

\begin{document}

\maketitle
{\renewcommand{\thefootnote}{}\footnotetext{Code is available at \url{https://github.com/engineerJPark/DriveIK}.}}

\begin{abstract}
\label{abstract}
Counterfactual obstacle insertion exposes blind planning in representative Driving VLAs: their trajectories remain largely unchanged despite a vehicle appearing directly ahead. An empirically calibrated bias--variance analysis identifies how ego-predictable training examples admit a shortcut through ego status and commands. We address this failure with an inverse-kinematics-inspired architecture that anchors trajectory decoding on current and predicted future visual states.
First, a next visual state prediction objective that requires the LLM to predict the future visual scene provides dense visual supervision and suppresses shortcut paths.
Second, a separate Inverse Kinematics Network (a cross-attention-based conditional diffusion model) that takes only the current and future visual states as input is designed to suppress reliance on ego status and textual shortcuts during trajectory decoding.
Our 0.5B model achieves state-of-the-art trajectory planning performance on NAVSIM-v1, NAVSIM-v2, and nuScenes using cameras alone, without LiDAR or a BEV encoder. Against the matched OpenDriveVLA* baseline, it improves NAVSIM PDMS and EPDMS by $19.0$ and $20.4$ points, respectively.
Controlled ablations and intervention analyses support improved use of visual features, with future-state prediction sharpening hazard-specific responses.
\end{abstract}

\vspace{-1em}
\section{Introduction}
\label{sec:intro}
\vspace{-1em}

\begin{figure*}[t]
\centering
   \includegraphics[width=1.0\linewidth,height=0.72\textheight,keepaspectratio]{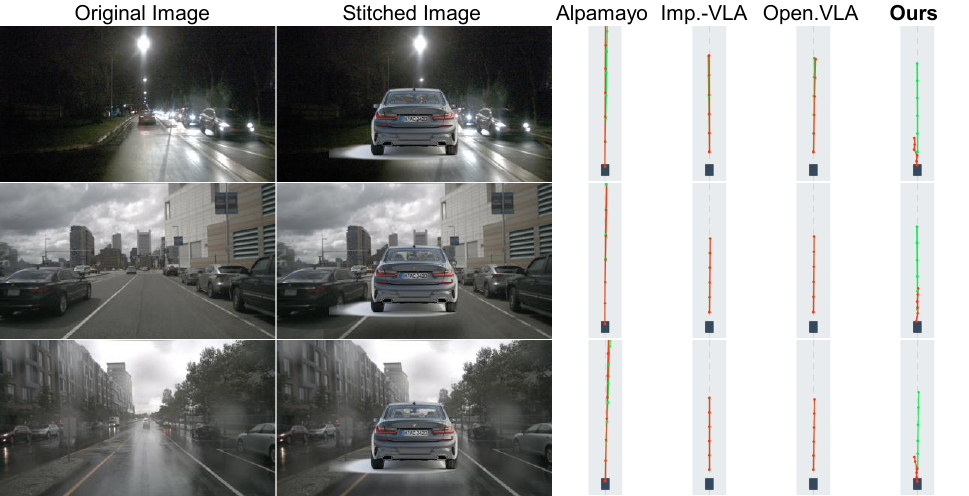}
\caption{Counterfactual obstacle stitching on nuScenes val. Left columns: original front-camera image and the same image with a near-range vehicle sprite stitched in front of the ego. Right columns: BEV trajectories before (green) and after (red) stitching for existing VLAs and our model. While existing VLAs' trajectory remains nearly invariant to the visual input, our model produces a distinct slowdown/avoidance reaction in response to the stitched obstacle, demonstrating that visual features causally influence its trajectory output.}
\label{fig:stitching}
\end{figure*}

In autonomous driving, Vision-Language-Action models (Driving VLA) are rapidly emerging as a unified policy that takes camera images and textual commands as input and directly outputs future driving trajectories \cite{hwang2024emma, zhou2025opendrivevla}.
Unlike conventional autonomous driving pipelines, which optimize perception, prediction, and planning as separate modules, this approach unifies the three stages into a single end-to-end policy within a VLM backbone, offering benefits in simplicity and scalability.

Camera conditioning alone does not ensure visually grounded planning. Our counterfactual experiments expose \emph{blind planning} in representative Driving VLAs: inserting a vehicle directly ahead produces little change in their predicted trajectories.
As shown in \cref{fig:stitching}, representative existing Driving VLAs \cite{zhou2025opendrivevla, wang2025alpamayo, chi2025impromptu} barely change their trajectory even when a vehicle is synthesized in front of the ego vehicle, maintaining the same straight path as before the synthesis, which exemplifies blind planning by definition.
Building on prior observations of ego-status shortcuts \cite{li2024ego}, we quantify this failure through controlled visual interventions that hold non-visual inputs fixed.

We identify a structural opportunity for blind planning in the coupling between trajectory supervision and decoder inputs.
Drawing an analogy to inverse kinematics -- where a trajectory is anchored by both a start and a target configuration
-- we observe that supplying a future visual state $V_{t+\Delta}$ alongside the current visual state $V_t$ provides the model an explicit reason to ground its trajectory in visual features.
In the direct-prediction formulation, the decoder can fit ego-predictable trajectories through ego status and commands without reconstructing the future visual scene. Our bias--variance analysis characterizes this shortcut under an empirically calibrated majority-regime assumption.
This is a typical shortcut-learning failure in end-to-end training~\cite{li2024ego}: when the visual scene is absent from the output space, the model is rarely rewarded for exploiting it.

\begin{figure*}[t]
\centering
   \includegraphics[width=1.0\linewidth,height=0.72\textheight,keepaspectratio]{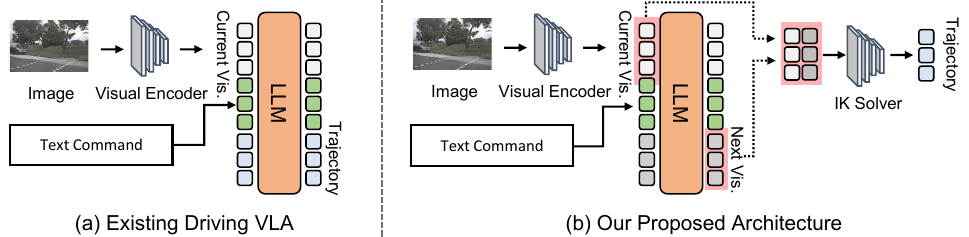}
\caption{Existing Driving VLA (a) vs.\ our model (b). Existing models decode the trajectory directly from the current visual tokens $V_t$, ego status $E$, and text command $C$. In contrast, ours requires the LLM to predict a future visual state $\hat V_{t+\Delta}$ and decodes the trajectory through an IK Network conditioned only on the pair $(V_t, \hat V_{t+\Delta})$, so that ego status and text commands do not directly enter the trajectory decoder.}
\label{fig:overview}
\end{figure*}

This points to two complementary gaps:
on the supervision side, the model is never asked to reason about future visual scenes,
and on the architecture side, the trajectory decoder is free to bypass visual tokens entirely and rely on ego status and text commands.
We address both at once, as illustrated in \cref{fig:overview}(b).
First, we separate the trajectory decoder into a dedicated Inverse Kinematics Network (IK Network),
a bottleneck conditioned only on the pair of the current visual token $V_t$ and the future visual token $\hat{V}_{t+\Delta}$ predicted by the backbone.
This removes the direct ego-status and command pathway to trajectory decoding and makes the current/predicted visual-state pair the action expert's conditioning interface.
We use the term ``inverse kinematics'' in a stylistic sense, following its spirit rather than the strict vehicle-dynamics formulation: a trajectory is regressed from the observed change between two visual states.
Second, for this bottleneck to be meaningful, $\hat{V}_{t+\Delta}$ must be explicitly produced by the backbone,
so we additionally introduce a next-state prediction objective that places the visual scene back into the output space, addressing the supervision-side gap.
Through these architectural and training modifications, the primary target of this paper is the recovery of visual grounding in Driving VLA.

To this end, we conduct a series of validations including counterfactual obstacle stitching, GradCAM, and per-sample statistical analyses, demonstrating both quantitatively and qualitatively that the model indeed comes to rely on visual features when determining trajectories.
Our 0.5B model achieves state-of-the-art trajectory planning performance on NAVSIM-v1, NAVSIM-v2, and nuScenes. It improves over the matched OpenDriveVLA* baseline by $19.0$ PDMS and $20.4$ EPDMS on NAVSIM while using cameras alone, without LiDAR or a BEV encoder.
More broadly, these results hint at a remedy for a common pitfall in end-to-end learning:
models tend to ignore any input modality that is not part of the supervision target.

Our contributions are summarized as follows.
\begin{itemize}
    \item \textbf{Theoretical Analysis}: We demonstrate via bias–variance decomposition that \emph{blind planning} is a theoretically derived loss-optimal shortcut rather than a training failure, caused by visual-feature penalization in standard trajectory losses.
    \item \textbf{Architectural Remedies}: To restore visual grounding, we introduce a next-state prediction objective and an IK-Network bottleneck that forces the model to process ego-status through visual features.
    \item \textbf{Superior Efficiency}: Through counterfactual obstacle stitching, and per-sample statistical analyses, the proposed design is shown to actually recover visual grounding, and as a result, achieves driving performance on NAVSIM v1, NAVSIM v2 and nuScenes comparable to VLAs more than an order of magnitude larger (7B--8B) using only a 0.5B model.
\end{itemize}

\vspace{-1em}
\section{Motivations}
\label{sec:motivations}
\vspace{-0.5em}

\subsection{Formulation of Existing Driving VLA Models}
\vspace{-0.2em}
Existing Driving VLA planners are trained to directly regress trajectories.
Let $V_t$ denote the current visual observation (e.g., images or BEV tokens), $E_t$ the ego status (velocity, yaw-rate, acceleration, etc.), $C_t$ a text command (route-level intent), and $T_t \in \mathbb{R}^{H \times d}$ the $H$-step future trajectory (waypoints or controls). The training objective is then defined as:
\begin{equation}
\hat T_t = f_\theta(V_t, E_t, C_t),
\qquad
\mathcal{L}_{\text{traj}}(\theta)
= \mathbb{E}\left[\lVert T_t - f_\theta(V_t,E_t,C_t)\rVert_2^2\right],
\label{eq:prior_direct_traj}
\end{equation}
or with text-based next token prediction
\begin{equation}
\hat T_t = f_\theta(V_t, E_t, C_t),
\qquad
\mathcal{L}_{\text{traj}}(\theta)
= \mathbb{E}\left[-\sum_{i=1}^{L}\log p_\theta\!\left(y_{t,i}^{\star}\mid y_{t,<i}^{\star},V_t,E_t,C_t\right)\right].
\label{eq:prior_token_traj}
\end{equation}
Directly regressing $\hat T_t=f_\theta(V_t,E_t,C_t)$ (Eq.~\eqref{eq:prior_direct_traj}) requires the VLA model to obtain a feasible control sequence without an explicit boundary condition.

\vspace{-0.2em}
\subsection{Phenomena due to Formulation of Existing VLA}
\vspace{-0.2em}
\noindent\textbf{Driving VLAs Ignore Visual Information.}
When deciding the next action in a driving situation, ego status, command, and visual information all serve as factors.
To characterize the competition between ego-predictable behavior and vision-dependent corrections, we decompose the target trajectory into two regimes. In the majority regime ($\mathbb{I}_t=0$), we assume that the short-horizon target is conditionally independent of vision given ego status and command; in the vision-critical regime, visual information supplies a trajectory correction. \Cref{sec:formulation} gives the detailed derivation and empirical calibration.
\begin{equation}
T_t
= T_0(E_t,C_t) \;+\; \mathbb{I}_t \cdot \Delta(V_t,E_t,C_t) \;+\; \xi_t,
\label{eq:traj_decompose}
\end{equation}
where $T_0$ is the dominant behavior driven by ego status and command (e.g., lane-following, cruising, stopping), $\Delta$ is an action correction conditioned on visual features (e.g., obstacle avoidance, yield timing, intersection choice), and $\xi_t$ is the noise inherent in the data. $\mathbb{I}_t \in \{0,1\}$ indicates whether the current situation is vision-critical or not.

Substituting the target decomposition (Eq.~\ref{eq:traj_decompose}) for $T_t$ and letting $\varepsilon\triangleq\mathbb{P}(\mathbb{I}_t=1)$, the loss function for trajectory prediction can be decomposed as follows:
\begin{equation}
\mathcal{L}(\theta)
=(1-\varepsilon)\,\mathcal{L}_0(\theta)+\varepsilon\,\mathcal{L}_1(\theta),
\label{eq:loss_mixture}
\end{equation}
where $\mathcal{L}_0$ and $\mathcal{L}_1$ denote the conditional losses under $\mathbb{I}_t=0$ and $\mathbb{I}_t=1$, respectively.

For each $(E,C)$, we can write the predictor in non-vision-critical cases $\bar f_\theta(E,C)$ and the $V$-conditional residual $\tilde f_\theta(V,E,C)$ as:
\begin{equation}
\bar f_\theta(E,C)\;\triangleq\;\mathbb{E}\left[f_\theta(V,E,C)\mid E,C,\mathbb{I}=0\right],
\quad
\tilde f_\theta(V,E,C)\;\triangleq\;f_\theta(V,E,C)-\bar f_\theta(E,C).
\label{eq:function_decomp}
\end{equation}
By definition, $\mathbb{E}[\tilde f_\theta(V,E,C)\mid E,C,\mathbb{I}=0]=0$.
Under $(E,C,\mathbb{I}=0)$, the target trajectory is $T_0(E,C)+\xi$ and does not depend on the visual feature $V$, so the conditional MSE under $\mathbb{I}=0$ can be decomposed in bias--variance form as follows (see \cref{sec:formulation_direct} for the detailed derivation):
\begin{equation}
\begin{aligned}
\mathcal{L}_0(\theta)
&= \left\|T_0(E,C)-\bar f_\theta(E,C)\right\|_2^2 \\
&\quad + \mathbb{E}\!\left[\left\|\tilde f_\theta(V,E,C)\right\|_2^2 \,\big|\, E,C,\mathbb{I}{=}0\right]
+ \mathbb{E}\!\left[\|\xi\|_2^2\right].
\end{aligned}
\label{eq:var_penalty}
\end{equation}
The first term is the bias corresponding to the average behavior that can be fit using only $(E,C)$, and the second term is the component whose output varies with $V$ entering directly as a loss penalty in the majority regime.
That is, the majority data weighted by $1-\varepsilon$ pushes the learning signal toward $\tilde f_\theta\to 0$, causing the model to ignore visual features.
Returning to the mixture loss \eqref{eq:loss_mixture}, the learning signal that drives the model to use visual features comes only from the minority term weighted by $\varepsilon$, but in real driving data $\varepsilon\ll 1$, making it structurally difficult to learn vision-grounded planning.
From this, we can interpret that shortcut learning through ego status and text command alone arises.

\vspace{-0.5em}
\subsubsection{An Architectural Remedy without Additional Data.}
\vspace{-0.5em}
We address the two shortcut pathways together: next-state prediction supplies dense visual supervision, and the IK Network removes direct ego-status and command access from trajectory decoding.
Our key concept is to define the current visual feature as the current state and let the VLA predict the next state.
Then, using the current/next states, trajectory prediction is performed through a separate inverse-kinematics module (IK module).
Here, $V_t$ denotes the 2D image feature obtained from the vision encoder.
The LLM decoder $w_\phi$ of the VLA and the IK module $h_\psi$ predict the next state $V_{t+\Delta}$ and the trajectory $T_t$, respectively.
$V_{t+\Delta}$ serves as a proxy terminal condition and contributes to forming the IK solution of the VLA.
\begin{equation}
\label{eq:ours_factorization}
\hat V_{t+\Delta} = w_\phi(V_t, E_t, C_t),
\qquad
\hat T_t = h_\psi(V_t, \hat V_{t+\Delta}).
\end{equation}
To this end, the two objectives are jointly trained.
\begin{equation}
\mathcal{L}(\phi,\psi)
=
\lambda_{\text{state}}\,
\mathbb{E}\left[\lVert V_{t+\Delta} - w_\phi(V_t,E_t,C_t)\rVert_2^2\right]
\;+\;
\lambda_{\text{traj}}\,
\mathbb{E}\left[\lVert T_t - h_\psi(V_t,\hat V_{t+\Delta})\rVert_2^2\right].
\label{eq:ours_loss}
\end{equation}
With this loss correction, concretely, the gradient with respect to $\theta_V$ is computed as:
\begin{equation}
\nabla_{\theta_V}\mathcal{L}
=
\lambda_{\text{state}}\nabla_{\theta_V}\mathcal{L}_{\text{state}}
+\lambda_{\text{traj}}\nabla_{\theta_V}\mathcal{L}_{\text{traj}},
\label{eq:grad_ours_sum}
\end{equation}
where $\nabla_{\theta_V}\mathcal{L}_{\text{state}}$ does not vanish even in the $\mathbb{I}_t=0$ situations that constitute most of the dataset, so that the visual parameters $\theta_V$ continue to receive a learning signal.
The new loss obtained by this task formulation correction is intended to suppress shortcut learning induced by ego status and command.
(i) Next state prediction provides dense supervision through visual features even in $\mathbb{I}_t=0$ situations.
(ii) The IK module $h_\psi$ performs trajectory prediction using only the current/next state $(V_t,\hat V_{t+\Delta})$ information, without explicit ego status or command feature, which discourages the optimization from collapsing into a direct shortcut mapping $(E_t,C_t)\rightarrow T_t$.

\vspace{-0.5em}
\section{Methods}
\label{sec:methods}
\vspace{-0.5em}

\begin{figure*}[t]
\centering
   \includegraphics[width=0.95\linewidth,height=0.72\textheight,keepaspectratio]{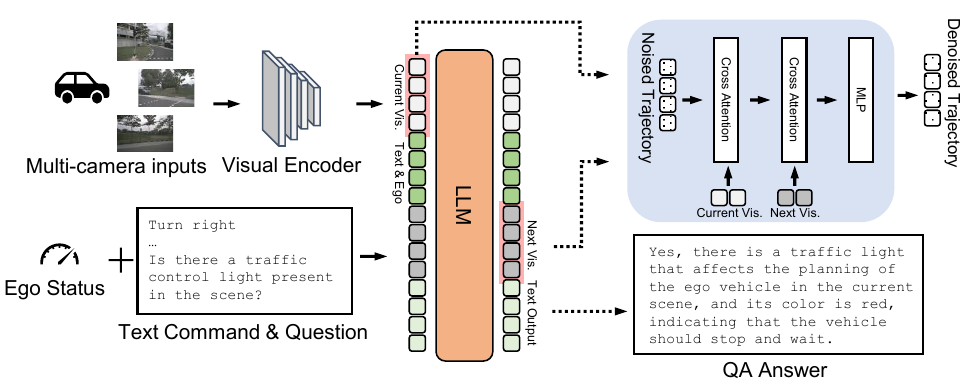}
   \caption{Overview of our model. The vision encoder extracts visual tokens $V_t$ from the three front cameras; the LLM consumes $(V_t, E, C)$ together with the QA prompt and is trained to predict the future visual state $\hat V_{t+\Delta}$ (and to answer driving-scene questions). The Inverse Kinematics Network, a cross-attention conditional diffusion model, decodes the trajectory from only the pair $(V_t, \hat V_{t+\Delta})$, without direct access to ego status $E$ or text command $C$.}
\label{fig:methods}
\end{figure*}

\subsection{Overview}
\vspace{-0.5em}
Our VLA consists of three mechanisms: Next Visual Token Prediction, Question and Answering, and the Inverse Kinematics Network.
We extract visual features by feeding front, front-left, and front-right images into a DINOv3 ViT-S/16 \cite{simeoni2025dinov3} encoder, and pass them through an MLP projector before inputting them to the LLM.
The LLM also receives ego status and text instructions encoded through an MLP, as well as questions about the driving scene.
As output, it predicts the next visual state at $N$ seconds ahead, and, when given a question, outputs the answer as text.
The current visual state and the next visual state are then fed into the Inverse Kinematics Network to perform trajectory prediction.
The detailed architecture of our VLA is shown in \cref{fig:methods}.

\vspace{-0.5em}
\subsection{Next Visual State Prediction}
\vspace{-0.5em}
As its first task, our VLA performs `Next Visual State' prediction.
We extract visual features by feeding front, front-left, and front-right images into a DINOv3 ViT-S/16 \cite{simeoni2025dinov3} encoder, and pass them through an MLP projector before inputting them to the LLM.
When fed into the LLM, the features are placed between \texttt{\textless{}scene\_start\textgreater{}} and \texttt{\textless{}scene\_end\textgreater{}}, and represented in text form by the placeholder \texttt{\textless{}SCENE\textgreater{}}.
Unlike prior work \cite{zhou2025opendrivevla}, we do not directly emit a trajectory text output here, but instead let the model predict the next visual state at $N$ seconds ahead.
This prediction is trained to be formatted between \texttt{\textless{}next\_start\textgreater{}} and \texttt{\textless{}next\_end\textgreater{}} in the same format as the input.
Concretely, we read out the LLM hidden states at the predicted \texttt{\textless{}next\_start\textgreater{}}\,...\,\texttt{\textless{}next\_end\textgreater{}} positions in the assistant turn,
 and supervise them with an MSE loss against the ground-truth future visual tokens obtained by passing the next-frame images through the same vision encoder and projector.
Writing $H_t$ and $A_{t-1}$ for the supplied trajectory history and previous-action context, respectively, the predictor and its visual supervision are
\[
\hat V_{t+\Delta}=w_\phi(V_t,E_t,C_t,H_t,A_{t-1}),
\qquad
\mathcal{L}_{\mathrm{visual}}=\operatorname{MSE}(\hat V_{t+\Delta},V_{t+\Delta}).
\]
The LLM fuses visual evidence with ego kinematics and the textual route command: $C_t$ specifies the intended maneuver, while $E_t$ supplies motion context for predicting a compatible future state.
The LLM encodes maneuver intent and ego kinematics in $\hat V_{t+\Delta}$, allowing language to guide trajectory generation without direct text input to the IK decoder.
See Appendix~\ref{sec:appendix_arch} for the predictor inputs and the IK conditioning interface.

\vspace{-0.5em}
\subsection{Inverse Kinematics Network}
\vspace{-0.5em}
As mentioned earlier, our method does not perform trajectory prediction directly within the LLM.
We design an Inverse Kinematics Network (IK network), which takes only the current visual state seen earlier and the next visual state predicted by the LLM as input, and predicts the next trajectory.
We stop trajectory-loss gradients at both conditioning inputs, separating visual-state learning from diffusion trajectory learning (\cref{sec:ik_gradient_isolation}).
Here, the constraint on the input is the key.
The IK Network does not take ego status $E$ or text command $C$ as input, and is conditioned only on $(V_t, \hat V_{t+\Delta})$.
This design choice distinguishes our work from existing world-model + policy combinations (e.g., DriveVLA-W0~\cite{li2025drivevla}, LAW~\cite{li2024enhancing}), in which future prediction is used as an auxiliary signal while the trajectory decoder still has access to the full $(E, C, V_t)$.
That is, our IK Network does not place the future visual state as an optional auxiliary input but uses the pair of two visual states as its primary condition, thereby suppressing the path along which an ego-status shortcut could otherwise flow.

As shown in \cref{fig:methods}, the IK network is constructed as a cross-attention-based conditional diffusion model.
Specifically, we stack two cross-attention layers in which a noisy trajectory $x_t \in \mathbb{R}^{2 \times 6}$ serves as the query and the current visual state token $V_t \in \mathbb{R}^{N \times H}$ and the next visual state token $\hat V_{t+\Delta} \in \mathbb{R}^{N \times H}$ predicted by the LLM serve as keys and values, predicting the noise $\epsilon$.
The timestep $t$ is added to the query through a sinusoidal embedding.
The denoising network $\epsilon_\theta$ is then defined as:
$$\epsilon_\theta(x_t, t, V_t, \hat V_{t+\Delta}) = \text{MLP}\Big(\text{CrossAttn}\big(\text{CrossAttn}(x_t + \text{TE}(t),~V_t),~\hat V_{t+\Delta}\big)\Big)$$
This model is trained with a DDPM~\cite{ho2020denoising} objective using the cosine $\beta$-schedule~\cite{nichol2021improved}.
$$\mathcal{L}_{ik} = \mathbb{E}_{t, x_0, \epsilon}\Big[\big\|\epsilon - \epsilon_\theta(\sqrt{\bar{\alpha}_t} x_0 + \sqrt{1-\bar{\alpha}_t}\,\epsilon,~t,~V_t,~\hat V_{t+\Delta})\big\|_2^2\Big]$$
At inference, we start from $x_T \sim \mathcal{N}(0, I)$ and run $50$ steps of DDPM sampling to obtain the final trajectory $T \in \mathbb{R}^{2 \times 6}$.

The IK Network turns the predicted visual state into an explicit input to action generation. Together with NSP supervision and gradient isolation, this interface couples trajectory decoding to a learned future scene while preserving the command and kinematic information needed to execute the maneuver.

\vspace{-0.5em}
\subsection{Question and Answering}
\vspace{-0.5em}
In parallel with the LLM's next visual state prediction, we train question and answering using scene description datasets such as nuCaption \cite{yang2025lidar} and NuX \cite{ding2024hint}.
By conducting this jointly with next visual state prediction, we enable the LLM to perform question and answering about the driving scene while simultaneously improving its scene understanding ability.
QA supervision strengthens the shared representation used for future-state prediction; see Appendix~\ref{sec:qa_planning_ablation} for the planning ablation.
When a question is fed into the LLM, it is placed between \texttt{\textless{}question\_start\textgreater{}} and \texttt{\textless{}question\_end\textgreater{}}, and the answer is trained to appear between \texttt{\textless{}answer\_start\textgreater{}} and \texttt{\textless{}answer\_end\textgreater{}} in the output.
This part is trained with next token prediction, identically to prior work \cite{zhou2025opendrivevla}.
$$\mathcal{L}(\theta) = - \frac{1}{T} \sum_{t=1}^{T} \log P_\theta(x_t \mid x_1, x_2, \dots, x_{t-1})$$

\vspace{-1em}
\section{Experiments}
\label{sec:experiments}
\vspace{-0.5em}

\subsection{Settings}
\label{sec:settings}
\vspace{-0.5em}

\noindent\textbf{Datasets and benchmarks.}
We train and evaluate our model on two types of autonomous driving benchmarks.
NAVSIM \cite{dauner2024navsim} is a challenging closed-loop benchmark, and we evaluate on the \texttt{NAVSIM-v1} and \texttt{NAVSIM-v2} splits using PDMS and EPDMS, respectively, along with detailed metrics such as NC, DAC, and TTC.
nuScenes \cite{caesar2020nuscenes} is a standard open-loop benchmark consisting of 700 (train) / 150 (val) scenes, where trajectory planning is evaluated by Average Displacement Error (ADE) and Collision Rate, and scene-level understanding is evaluated on nu-Caption \cite{yang2025lidar}, nu-X \cite{ding2024hint} and nuScenes-QA \cite{qian2024nuscenes}.

\noindent\textbf{Architecture.}
We use Qwen2.5-0.5B-Instruct \cite{yang2024qwen25} as the LLM backbone and DINOv3 ViT-S/16 \cite{simeoni2025dinov3} as the visual encoder for both datasets, since no pretrained BEV encoder is publicly available for NAVSIM.
The IK Network is a cross-attention-based conditional diffusion model trained with a cosine $\beta$-schedule \cite{nichol2021improved} and run for $50$ DDPM \cite{ho2020denoising} sampling steps at inference. 
Details are provided in \cref{sec:appendix_arch}.

\noindent\textbf{Training Setup.}
On NAVSIM, we follow the 2-stage SFT $\rightarrow$ GRPO \cite{shao2024deepseekmath} procedure of RecogDrive \cite{li2025recogdrive} on the navtrain split, with a $4$\,s future frames and $8$ output waypoints.
On nuScenes, we train for $4$ epochs jointly with nuScenes-QA, nu-Caption, and nu-X, using a $3$\,s frames and $6$ waypoints.
Details are listed in \cref{sec:appendix_training}.

\noindent\textbf{Baselines.}
On nuScenes, we compare against recent Driving VLAs \cite{hwang2024emma, tian2024drivevlm, wang2025alpamayo, chi2025impromptu}, such as OpenDriveVLA \cite{zhou2025opendrivevla}.
On NAVSIM, in addition to a NAVSIM-adapted variant of OpenDriveVLA \cite{zhou2025opendrivevla}, we include VLA models spanning 2B--8B backbones \cite{feng2025artemis, li2025drivevla, li2025recogdrive, luo2025adathinkdrive, zhou2025autovla, dang2026drivefine}.
Since the original OpenDriveVLA relies on a UniAD BEV encoder trained on nuScenes but no pretrained BEV encoder corresponding to NAVSIM is publicly available, we reimplement OpenDriveVLA for NAVSIM (denoted \textit{OpenDriveVLA*} in the tables) by unifying the visual encoder with DINOv3 \cite{simeoni2025dinov3} as in our model and generating the trajectory as a direct text output of the LLM as in the original paper.
This baseline shares with our model the visual encoder, LLM backbone (0.5B), training data, optimizer, learning rate schedule, warmup ratio, batch size, gradient clipping, encoder freeze policy, and the SFT $\rightarrow$ GRPO 2-stage training procedure used on NAVSIM, and the only difference between the two models is the introduction of the next visual state prediction objective and the IK Network.
This matched comparison evaluates the combined NSP and IK design while controlling backbone capacity, data, and training schedule; the conditioning ablations further examine the individual contributions.
Our model uses three perspective cameras without LiDAR or a BEV encoder, whereas DiffusionDrive and TransFuser use camera--LiDAR inputs~\cite{liao2025diffusiondrive,chitta2022transfuser}.
Sensor configurations are listed in the benchmark tables, with further comparison details in Appendix~\ref{sec:ablation_study}.

\vspace{-0.5em}
\subsection{Main Experiments}
\vspace{-0.5em}

\subsubsection{Experiments on NAVSIM.}
\label{sec:main_experiments_navsim}
\begin{table}[t]
\caption{PDMS results on the \texttt{NAVSIM-v1} benchmark. \textbf{Bold}=best, \underline{underline}=second best. $\dagger$ denotes models trained with the SFT$\rightarrow$GRPO schedule. OpenDriveVLA* shares all training hyperparameters, data, and the SFT$\rightarrow$GRPO schedule with our model; the only differences are the next visual state prediction objective and the IK Network.}

\centering
\setlength{\tabcolsep}{3pt}
\resizebox{\linewidth}{!}{%
\renewcommand{\arraystretch}{0.9}
\begin{tabular}{lcccccccc}
\toprule
\textbf{Method} & \textbf{Size} & \textbf{Sensors} & \textbf{NC}$\uparrow$ & \textbf{DAC}$\uparrow$ & \textbf{TTC}$\uparrow$ & \textbf{C.}$\uparrow$ & \textbf{EP}$\uparrow$ & \textbf{PDMS}$\uparrow$ \\
\midrule
Human & - & -- & 100 & 100 & 100 & 99.9 & 87.5 & 94.8 \\
\midrule
\multicolumn{9}{c}{\textit{End-to-End Methods}} \\
\midrule
UniAD \cite{hu2023planning}            & -     & C   & 97.8                & 91.9                & 92.9                & \textbf{100.0}      & 78.8                & 83.4 \\
TransFuser \cite{chitta2022transfuser} & -     & C+L & 97.7                & 92.8                & 92.8                & \textbf{100.0}      & 79.2                & 84.0 \\
LAW \cite{li2024enhancing}             & -     & C   & 96.4                & 95.4                & 88.7                & \underline{99.9}    & 81.7                & 84.6 \\
Hydra-MDP \cite{li2024hydra}           & -     & C+L & 98.3                & 96.0                & 94.6                & \textbf{100.0}      & 78.7                & 86.5 \\
DiffusionDrive \cite{liao2025diffusiondrive} & -     & C+L & 98.2          & 96.2                & 94.7                & \textbf{100.0}      & 82.2                & 88.1 \\
WoTE \cite{li2025end}                  & -     & C+L & 98.5                & 96.8                & 94.4                & \underline{99.9}    & 81.9                & 88.3 \\
SUPER-AD \cite{ryu2025superad}         & -     & C   & 98.1                & 97.0                & 93.9                & 99.8                & 81.5                & 87.7 \\
\midrule
\multicolumn{9}{c}{\textit{Vision Language Action Methods}} \\
\midrule
AutoVLA \cite{zhou2025autovla}         & 3B  & C   & 98.4                & 95.6                & 98.0                & \underline{99.9}    & 81.9                & 89.1 \\
DriveVLA-W0 \cite{li2025drivevla}      & 7B  & C   & 98.7                & \textbf{99.1}       & 95.3                & 99.3                & 83.3                & 90.2 \\
AdaThinkDrive \cite{luo2025adathinkdrive} & 2B & C & 98.4               & 97.8                & 95.2                & \textbf{100.0}      & 84.4                & 90.3 \\
ReCogDrive \cite{li2025recogdrive}     & 2B  & C   & 98.2                & 97.8                & 95.2                & 99.8                & 83.5                & 89.6 \\
DriveFine \cite{dang2026drivefine}     & 8B  & C   & \underline{98.8}    & \underline{98.6}    & 96.2                & \textbf{100.0}      & \textbf{86.9}       & \underline{91.8} \\
\midrule
OpenDriveVLA* \cite{zhou2025opendrivevla} & 0.5B & C & 92.2               & 83.2                & 91.2                & \underline{99.9}    & 68.7                & 73.2 \\
Ours (SFT)                             & 0.5B & C   & \textbf{99.1}{\tiny\inc{(+6.9)}}  & 96.2{\tiny\inc{(+13.0)}} & \underline{99.0}{\tiny\inc{(+7.8)}} & \textbf{100.0}{\tiny\inc{(+0.1)}} & 83.4{\tiny\inc{(+14.7)}} & 90.3{\tiny\inc{(+17.1)}} \\
Ours$^{\dagger}$                       & 0.5B & C   & \textbf{99.1}{\tiny\inc{(+6.9)}}  & 97.8{\tiny\inc{(+14.6)}} & \textbf{99.1}{\tiny\inc{(+7.9)}} & \underline{99.9}{\tiny\eqv{(+0.0)}} & \underline{85.4}{\tiny\inc{(+16.7)}} & \textbf{92.2}{\tiny\inc{(+19.0)}} \\
\bottomrule
\end{tabular}
\label{tab:navsim_v1_exp}
}
\par\smallskip
{\raggedright\scriptsize Sensors: C = cameras; L = LiDAR; C+L = cameras and LiDAR. Entries describe inference-time scene sensors, not ego status or training supervision.\par}
\end{table}
\begin{table}[t]
\caption{EPDMS results on the \texttt{NAVSIM-v2} benchmark. \textbf{Bold}=best, \underline{underline}=second best.}

\centering
\setlength{\tabcolsep}{3pt}
\resizebox{\linewidth}{!}{%
\begin{tabular}{lcccccccccccc}
\toprule
\textbf{Method} & \textbf{Size} & \textbf{Sensors} & \textbf{NC}$\uparrow$ & \textbf{DAC}$\uparrow$ & \textbf{DDC}$\uparrow$ & \textbf{TLC}$\uparrow$ & \textbf{EP}$\uparrow$ & \textbf{TTC}$\uparrow$ & \textbf{LK}$\uparrow$ & \textbf{C.}$\uparrow$ & \textbf{EC}$\uparrow$ & \textbf{EPDMS}$\uparrow$ \\
\midrule
Ego-MLP \cite{dauner2024navsim}              & -     & -- & 93.1                & 77.9                & 92.7                & 99.6                & 86.0                & 91.5                & 89.4                & \underline{98.3}    & 85.4                & 64.0 \\
TransFuser \cite{chitta2022transfuser}        & -     & C+L & 96.9                & 89.9                & 97.8                & 99.7                & 87.1                & 95.4                & 92.7                & \underline{98.3}    & \underline{87.2}    & 76.7 \\
DriveSuprim \cite{yao2026drivesuprim}         & -     & C & 97.5                & 96.5                & \underline{99.4}    & 99.6                & 88.4                & 96.6                & 95.5                & \underline{98.3}    & 77.0                & 83.1 \\
ARTEMIS \cite{feng2025artemis}                & -- & C+L & 98.3                & 95.1                & 98.6                & \underline{99.8}    & 81.5                & 97.4                & 96.5                & \underline{98.3}    & --                  & 83.1 \\
ReCogDrive \cite{li2025recogdrive}            & 2B    & C & 98.3                & 95.2                & \textbf{99.5}       & \underline{99.8}    & 87.1                & 97.5                & 96.6                & \underline{98.3}    & 86.5                & 83.6 \\
DiffusionDrive \cite{liao2025diffusiondrive}  & -     & C+L & 98.2                & 95.9                & \underline{99.4}    & \underline{99.8}    & 87.5                & 97.3                & 96.8                & \underline{98.3}    & \textbf{87.7}       & 84.5 \\
SUPER-AD \cite{ryu2025superad}                & -     & C & 98.1                & 97.3                & 98.7                & \underline{99.8}    & 86.8                & 97.3                & \underline{97.5}    & \underline{98.3}    & 86.1                & 84.3 \\
DriveVLA-W0 \cite{li2025drivevla}             & 7B    & C & 99.0                & \textbf{98.4}       & 99.3                & \textbf{99.9}       & 87.0                & 98.1                & 93.2                & 97.9                & 58.9                & 86.5 \\
DriveFine \cite{dang2026drivefine}            & 8B    & C & 98.7                & 97.3                & \textbf{99.5}       & \underline{99.8}    & \underline{88.7}    & 97.8                & \textbf{97.7}       & \textbf{98.4}       & 83.8                & \underline{89.7} \\
\midrule
OpenDriveVLA* \cite{zhou2025opendrivevla}     & 0.5B  & C & 92.2                & 83.2                & 93.8                & 99.4                & \textbf{89.3}       & 91.2                & 90.8                & \underline{98.3}    & 76.8                & 70.2 \\
Ours (SFT)                                    & 0.5B  & C & \underline{99.1}{\tiny\inc{(+6.9)}} & 96.2{\tiny\inc{(+13.0)}} & \textbf{99.5}{\tiny\inc{(+5.7)}} & \textbf{99.9}{\tiny\inc{(+0.5)}} & 87.7{\tiny\dec{(-1.6)}} & \underline{99.1}{\tiny\inc{(+7.9)}} & \underline{97.5}{\tiny\inc{(+6.7)}} & 98.2{\tiny\dec{(-0.1)}} & 84.4{\tiny\inc{(+7.6)}} & 89.3{\tiny\inc{(+19.1)}} \\
Ours$^{\dagger}$                              & 0.5B  & C & \textbf{99.2}{\tiny\inc{(+7.0)}} & \underline{97.8}{\tiny\inc{(+14.6)}} & \textbf{99.5}{\tiny\inc{(+5.7)}} & \textbf{99.9}{\tiny\inc{(+0.5)}} & 88.4{\tiny\dec{(-0.9)}} & \textbf{99.2}{\tiny\inc{(+8.0)}} & 97.4{\tiny\inc{(+6.6)}} & 98.2{\tiny\dec{(-0.1)}} & 81.3{\tiny\inc{(+4.5)}} & \textbf{90.6}{\tiny\inc{(+20.4)}} \\
\bottomrule
\end{tabular}
\label{tab:navsim_v2_exp}
}
\par\smallskip
{\raggedright\scriptsize Sensors: C = cameras; L = LiDAR; C+L = cameras and LiDAR; -- = no scene-sensor input for Ego-MLP. Sensors describe inference inputs, not training supervision. ARTEMIS uses a ResNet-34-based sensor encoder; its parameter count is left unspecified. $\dagger$: SFT$\rightarrow$GRPO.\par}
\end{table}

\vspace{-0.5em}

As shown in \cref{tab:navsim_v1_exp}, on NAVSIM-v1 our 0.5B model achieves the highest PDMS of $92.2$, an improvement of $+19.0$ over the matched OpenDriveVLA* baseline ($73.2$) with the same backbone.
This score exceeds those of the larger VLA baselines, including DriveVLA-W0 \cite{li2025drivevla} (7B), ReCogDrive \cite{li2025recogdrive} (2B), AdaThinkDrive \cite{luo2025adathinkdrive} (2B), and AutoVLA \cite{zhou2025autovla} (3B), and surpasses the 8B-scale DriveFine \cite{dang2026drivefine} ($91.8$) by $0.4$ points.
In terms of detailed metrics, our model also records No-Collision (NC) $99.1$ and Time-To-Collision (TTC) $99.1$, demonstrating strong performance on collision avoidance and time-to-collision in addition to overall PDMS.

As shown in \cref{tab:navsim_v2_exp}, on the more challenging NAVSIM-v2 benchmark, our 0.5B model achieves the highest EPDMS of $90.6$, exceeding the $16\times$ larger DriveFine-8B ($89.7$) by $0.9$ points and DriveVLA-W0-7B ($86.5$) by $4.1$ points. The gain over the matched OpenDriveVLA* baseline ($70.2$) is $20.4$ points.
Our model also achieves the highest NC ($99.2$) and TTC ($99.2$), and ties the highest TLC ($99.9$), demonstrating strong collision avoidance and traffic-light compliance. Lane Keeping reaches $97.4$.

These results have two implications.
First, since the only difference from the OpenDriveVLA* baseline is our next visual state prediction and IK Network, the $+19.0$ PDMS and $+20.4$ EPDMS gains support the combined design under the matched configuration.
Second, the fact that a 0.5B light model has reached driving performance on par with 7B--8B VLAs suggests that structural changes that force active use of visual features may have a greater impact on trajectory planning accuracy than simply scaling up the LLM.

\vspace{-0.5em}
\subsubsection{Experiments on nuScenes.}
\label{sec:main_experiments_nuscenes}
\vspace{-0.5em}
\begin{table*}[t]
\caption{Open-Loop planning performance on nuScenes \textit{val set}. ``-'' indicates the metric is not reported for that method in the paper's tables. \textbf{Bold}=best, \underline{Underline}=second.}
\centering
\setlength{\tabcolsep}{2pt}
\resizebox{\linewidth}{!}{%
\begin{tabular}{lcccccccc|cccccccc|ccc}
\toprule
\textbf{Method} & \multicolumn{8}{c}{\textbf{ST-P3 metrics}} & \multicolumn{8}{c}{\textbf{UniAD metrics}} & \textbf{LLM} & \textbf{Input} & \textbf{Sensors} \\
\cmidrule(lr){2-9} \cmidrule(lr){10-17}
 & \multicolumn{4}{c}{\textbf{L2 (m)} $\downarrow$} & \multicolumn{4}{c}{\textbf{Collision (\%)} $\downarrow$} & \multicolumn{4}{c}{\textbf{L2 (m)} $\downarrow$} & \multicolumn{4}{c}{\textbf{Collision (\%)} $\downarrow$} & & & \\
\cmidrule(lr){2-5} \cmidrule(lr){6-9} \cmidrule(lr){10-13} \cmidrule(lr){14-17}
 & 1s & 2s & 3s & Avg. & 1s & 2s & 3s & Avg. & 1s & 2s & 3s & Avg. & 1s & 2s & 3s & Avg. & & & \\
\midrule
\multicolumn{20}{c}{\textbf{Non-Autoregressive Methods}} \\
\midrule
ST-P3 \cite{hu2022st}           & 1.33 & 2.11 & 2.90 & 2.11 & 0.23 & 0.62 & 1.27 & 0.71 & - & - & - & - & - & - & - & - & - & Visual & C \\
VAD \cite{jiang2023vad}          & 0.17 & 0.34 & 0.60 & 0.37 & 0.07 & 0.10 & 0.24 & 0.14 & - & - & - & - & - & - & - & - & - & Visual & C \\
Ego-MLP \cite{zhai2023rethinking}& 0.46 & 0.76 & 1.12 & 0.78 & 0.21 & 0.35 & 0.58 & 0.38 & - & - & - & - & - & - & - & - & - & Ego & -- \\
UniAD \cite{hu2023planning}      & 0.44 & 0.67 & 0.96 & 0.69 & 0.04 & 0.08 & 0.23 & 0.12 & 0.48 & 0.96 & 1.65 & 1.03 & 0.05 & 0.17 & 0.71 & 0.31 & - & Visual & C \\
InsightDrive \cite{song2025insightdrive} & 0.23 & 0.41 & 0.68 & 0.44 & 0.09 & 0.10 & 0.27 & 0.15 & 0.30 & 0.72 & 1.41 & 0.81 & 0.08 & 0.15 & 0.84 & 0.36 & - & Visual & C \\
FF \cite{hu2021safe}             & - & - & - & - & - & - & - & - & 0.55 & 1.20 & 2.54 & 1.43 & 0.06 & 0.17 & 1.07 & 0.43 & - & LiDAR & L \\
EO \cite{khurana2022differentiable} & - & - & - & - & - & - & - & - & 0.67 & 1.36 & 2.78 & 1.60 & 0.04 & \underline{0.09} & 0.88 & 0.33 & - & LiDAR & L \\
\midrule
\multicolumn{20}{c}{\textbf{Autoregressive Methods}} \\
\midrule
GPVL \cite{li2025generative}     & 0.21 & 0.39 & 0.69 & 0.43 & 0.07 & 0.09 & 0.27 & 0.14 & - & - & - & - & - & - & - & - & BERT & Textual & C \\
DriveVLM \cite{tian2024drivevlm} & 0.18 & 0.34 & 0.68 & 0.40 & 0.10 & 0.22 & 0.45 & 0.27 & - & - & - & - & - & - & - & - & Qwen-VL-7B & Visual & C \\
GPT-Driver \cite{mao2023gpt}     & 0.20 & 0.40 & 0.70 & 0.44 & 0.04 & 0.12 & 0.36 & 0.17 & 0.27 & 0.74 & 1.52 & 0.84 & 0.07 & 0.15 & 1.10 & 0.44 & GPT-3.5 & Textual & -- \\
RDA-Driver \cite{huang2024making} & 0.17 & 0.37 & 0.69 & 0.40 & \underline{0.01} & \textbf{0.05} & 0.26 & \underline{0.10} & 0.23 & 0.73 & 1.54 & 0.80 & \textbf{0.00} & 0.13 & 0.83 & 0.32 & LLaVA-7B & Visual & C \\
OmniDrive \cite{wang2025omnidrive} & 0.14 & 0.29 & 0.55 & 0.33 & \textbf{0.00} & 0.13 & 0.78 & 0.30 & - & - & - & - & - & - & - & - & LLaVA-7B & Visual & C \\
EMMA \cite{hwang2024emma}        & 0.14 & 0.29 & 0.54 & 0.32 & - & - & - & - & - & - & - & - & - & - & - & - & Gemini & Visual & C \\
OpenEMMA \cite{xing2025openemma} & 1.45 & 3.21 & 3.76 & 2.81 & - & - & - & - & - & - & - & - & - & - & - & - & Qwen-VL-7B & Visual & C \\
DME-Driver \cite{han2025dme}     & - & - & - & - & - & - & - & - & 0.45 & 0.91 & 1.58 & 0.98 & 0.05 & 0.28 & \underline{0.55} & 0.29 & LLaVA-7B & Visual & C \\
OpenDriveVLA-0.5B \cite{zhou2025opendrivevla} & 0.15 & 0.32 & 0.57 & 0.35 & \underline{0.01} & \underline{0.06} & \underline{0.20} & \textbf{0.09} & 0.21 & 0.60 & 1.22 & 0.68 & \textbf{0.00} & 0.15 & 0.63 & 0.26 & Qwen2.5-0.5B & Visual & C \\
OpenDriveVLA-3B \cite{zhou2025opendrivevla}  & 0.14 & 0.30 & 0.55 & 0.33 & 0.02 & 0.07 & 0.22 & \underline{0.10} & \underline{0.19} & \underline{0.58} & 1.24 & 0.67 & \underline{0.02} & 0.18 & 0.70 & 0.30 & Qwen2.5-3B & Visual & C \\
OpenDriveVLA-7B \cite{zhou2025opendrivevla}  & 0.15 & 0.31 & 0.55 & 0.33 & \underline{0.01} & 0.08 & 0.21 & \underline{0.10} & 0.20 & \underline{0.58} & \underline{1.21} & \underline{0.66} & \textbf{0.00} & 0.22 & \underline{0.55} & \underline{0.25} & Qwen2.5-7B & Visual & C \\
Impromptu-VLA-3B \cite{chi2025impromptu} & \underline{0.13} & \underline{0.27} & \underline{0.52} & \underline{0.30} & - & - & - & - & - & - & - & - & - & - & - & - & Qwen2.5-VL-3B & Visual & C \\
Impromptu-VLA-7B \cite{chi2025impromptu} & \underline{0.13} & \underline{0.27} & 0.53 & \underline{0.30} & - & - & - & - & - & - & - & - & - & - & - & - & Qwen2.5-VL-7B & Visual & C \\
\midrule
Ours-0.5B & \textbf{0.04} & \textbf{0.06} & \textbf{0.10} & \textbf{0.06} & 0.07 & 0.08 & \textbf{0.11} & \textbf{0.09} & \textbf{0.05} & \textbf{0.08} & \textbf{0.25} & \textbf{0.13} & \underline{0.02} & \textbf{0.03} & \textbf{0.12} & \textbf{0.06} & Qwen2.5-0.5B & Visual & C \\
\bottomrule
\end{tabular}
\label{tab:main_exp}
}
\par\smallskip
{\raggedright\scriptsize Sensors: C = cameras; L = LiDAR; C+L = cameras and LiDAR. Sensors denote the scene-sensing modality, whereas Input denotes the planner's representation. Ego-MLP uses ego status only. GPT-Driver consumes textual perception/prediction outputs; -- leaves the upstream sensor/annotation configuration of the quoted result unspecified. Camera-based BEV encoders (e.g., OpenDriveVLA's UniAD encoder) are labeled C, not C+L. Ours uses perspective-camera features without a BEV encoder.\par}
\end{table*}

As shown in \cref{tab:main_exp}, our 0.5B model achieves an ADE of $0.06$, reducing the error by $-0.27$ relative to the $14\times$ larger 7B-scale OpenDriveVLA, and also records a collision rate of $0.09$ at a level equal to or above it.
Together with the component ablations below, these results show that the visual grounding introduced by next-state prediction and the IK Network translates into more accurate trajectory planning, enabling our 0.5B model to outperform the 7B-scale OpenDriveVLA with a $14\times$ smaller backbone.
As a side effect of the recovered visual grounding, our model also shows consistent improvements over OpenDriveVLA on the nu-Caption captioning and nu-X reasoning benchmarks.
Detailed results are reported in the appendix (\cref{sec:nucaption_appendix,sec:nux_appendix,sec:main_experiments_nuqa}).

\vspace{-0.5em}
\subsection{Analysis}
\label{sec:analysis}
\vspace{-0.5em}

\subsubsection{Counterfactual Stitching.}
\vspace{-0.5em}
\label{sec:stitching}
We first revisit the counterfactual probe introduced as the motivation in \cref{fig:stitching} of \cref{sec:intro}, where a near-range vehicle sprite is synthesized in front of the ego on nuScenes \emph{val} while all textual inputs (ego status, command, history) are held fixed.
Visually, our model is the only one that bends or shortens the trajectory in response to the stitched obstacle, while the baselines maintain a near-identical path before and after the perturbation.
\cref{tab:stitching_exp} quantifies this gap: our 0.5B model decelerates on $63.8\%$ of perturbed samples with an average $3$-second endpoint shortening of $-1.04$\,m, far exceeding OpenDriveVLA-0.5B ($44.2\%$, $-0.06$\,m) and even the much larger Alpamayo-R1-10B ($54.5\%$, $-0.12$\,m).
This suggests that, unlike existing Driving VLAs, our model relies on visual features when shaping the trajectory rather than treating them as a largely unused side input.

\begin{table*}[t]
\centering
\setlength{\tabcolsep}{3pt}
\begin{minipage}[t]{0.47\textwidth}
\centering
\caption{Obstacle stitching experiment.}
\label{tab:stitching_exp}
\vspace{1mm}
\resizebox{\linewidth}{!}{%
\begin{tabular}{lcc}
\toprule
\textbf{Model} & \textbf{Decel.\ Ratio} & \textbf{Avg.\ $\Delta$displ}  \\
\midrule
Impromptu-VLA 3B \cite{chi2025impromptu}      & 38.8\% & -0.00m \\
Alpamayo-R1 10B  \cite{wang2025alpamayo}      & 54.5\% & -0.12m \\
OpenDriveVLA 0.5B \cite{zhou2025opendrivevla} & 44.2\% & -0.06m \\
\midrule
Ours 0.5B & \textbf{63.8\%} & \textbf{-1.04m} \\
\bottomrule
\end{tabular}}
\end{minipage}\hfill
\begin{minipage}[t]{0.50\textwidth}
\centering
\caption{Conditioning ablation. L2: ST-P3 / UniAD. Hazard, Sky: deceleration rate (\%) with a sprite on the road / in the sky; Gap: Hazard$-$Sky (pp).}
\label{tab:ablation_study_main}
\vspace{1mm}
\resizebox{\linewidth}{!}{%
\begin{tabular}{lcccc}
\toprule
\textbf{IK conditioning} & \textbf{L2 (m)}$\downarrow$ & \textbf{Hazard} & \textbf{Sky} & \textbf{Gap}$\uparrow$ \\
\midrule
(0) None (K/V zeroed) & 5.58 / 8.62 & 0.0 & 0.0 & 0.0 \\
(0+) Ego/command only & 0.60 / 1.11 & 0.0 & 0.0 & 0.0 \\
(i) Current state only & 0.55 / 1.01 & 71.3 & 65.5 & 5.8 \\
(ii) $V_t+\hat V_{t+\Delta}$ (ours) & \textbf{0.53 / 0.98} & 82.8 & 45.2 & 37.6 \\
(iii) $V_t+V_{t+\Delta}^{\mathrm{GT}}$ & 2.58 / 4.05 & 95.4 & 46.1 & 49.3 \\
\bottomrule
\end{tabular}}
\end{minipage}
\end{table*}

Complementary GradCAM results and their measurement protocol are provided in Appendix~\ref{sec:gradcam_methodology}.

\begin{figure*}[t]
\centering
   \includegraphics[width=1.0\linewidth,height=0.72\textheight,keepaspectratio]{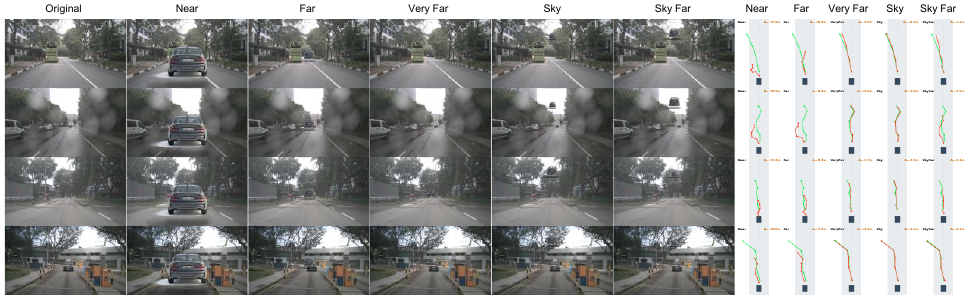}
\caption{Qualitative examples of the position--size stitching variants on nuScenes \emph{val}. For each scene, we visualize the predicted trajectory before (green) and after (red) stitching a vehicle sprite at four placements: Near and Far on the road, and Sky and SkyFar in the sky region (Far and SkyFar share the same sprite scale). The trajectory visibly bends or shortens for road-placed sprites, while sky-placed sprites of the same size leave the trajectory essentially unchanged, consistent with the quantitative results in \cref{tab:stitching_spatial}.}
\label{fig:stitching_position}
\end{figure*}

\begin{table}[htbp]
\centering
\caption{
Stitching response on nuScenes val split ($n=6019$), decomposed by sprite size and position, measured with paired same-seed sampling (K=50 DDIM).
$\Delta$ is the signed 3\,s endpoint displacement (after $-$ before, in meters); $|\Delta|$ is its magnitude.
Far and SkyFar share sprite scale ($0.5$) and differ only in position (road horizon vs.\ sky), forming the decisive paired control.
}
\label{tab:stitching_spatial}
\setlength{\tabcolsep}{4pt}
\resizebox{0.95\linewidth}{!}{%
\begin{tabular}{lcccccc}
\toprule
Stitching & scale & position
  & mean $\Delta$ (m) & mean $|\Delta|$ (m) & median $|\Delta|$ (m) & $|\Delta|<1$\,m (\%) \\
\midrule
Near (\cref{sec:stitching}) & $1.8$  & ego front (road) & $\mathbf{-1.04}$ & $2.51$ & $1.95$ & $27.8$ \\
Far                         & $0.5$  & road horizon     & $-0.54$ & $1.81$ & $1.20$ & $44.0$ \\
Very-Far                    & $0.25$ & horizon (small)  & $-0.09$ & $0.74$ & $0.43$ & $77.0$ \\
Sky                         & $0.30$ & sky (small)      & $-0.00$ & $0.72$ & $0.44$ & $77.6$ \\
Sky-Far                     & $0.5$  & sky              & $+0.10$ & $1.04$ & $0.70$ & $63.2$ \\
\midrule
\multicolumn{7}{l}{\emph{Far vs.\ SkyFar paired control} ($n=6019$, same scale $0.5$, position only):} \\
\multicolumn{7}{l}{\quad $|\Delta_{\text{Far}}| > |\Delta_{\text{SkyFar}}|$: $\mathbf{65.5\%}$\quad\quad signed paired mean $\Delta_{\text{Far}} - \Delta_{\text{SkyFar}} = -0.64$\,m} \\
\bottomrule
\end{tabular}
}
\end{table}

\vspace{-0.5em}
\subsubsection{Evaluating Position Awareness in the Stitching Response.}
\vspace{-0.5em}
\label{sec:stitching_spatial}

To separate whether the trajectory shortening reported in \cref{sec:stitching} originates from grounding on the positional meaning of the obstacle or merely from a response to the perceptual properties of the sprite,
we compare five stitching variants in which the position (road horizon vs.\ sky) and size (scale $0.25$\,$\sim$\,$1.8$) of the sprite are independently varied (\cref{tab:stitching_spatial}).
All measurements use paired same-seed sampling (the trajectories before and after stitching share the same DDIM noise), with $K=50$ samples averaged per token, so that $\Delta$ reflects the stitching-induced change rather than sampling stochasticity.

The signed mean $\Delta$ scales monotonically with both the realism of the obstacle and its size: Near (road, scale $1.8$) yields $-1.04$\,m,
Far (road horizon, scale $0.5$) yields $-0.54$\,m, while sprites placed in the sky (Sky, scale $0.30$; SkyFar, scale $0.50$) have signed means within $\pm 0.10$\,m and the sample-level reaction is small ($|\Delta|<1$\,m on $77.6\%$ and $63.2\%$ of scene frames, respectively).
The decisive test is the paired control between Far and SkyFar, which share the same sprite scale ($0.5$) and differ only in position: for the same frame, $\mathbf{65.5\%}$ of samples show a larger response when the sprite lies on the road than when it lies in the sky, with a signed paired mean of $\Delta_{\text{Far}} - \Delta_{\text{SkyFar}} = -0.64$\,m.
A conditional analysis on the $1{,}442$ scene frames for which \emph{Near} triggers a strong slowdown ($\Delta < -3$\,m) further supports this: under the same scene frames, the model essentially ignores the Sky and Very-Far variants ($|\Delta|<1$\,m on $76\%$ and $74\%$ of those tokens), while still reacting on Far ($|\Delta|<1$\,m on only $35\%$).

These results indicate that the response is genuinely position-aware rather than a generic size-based reaction:
when sprite size is held fixed, the model distinguishes between road and sky, and conditional on a strong on-road reaction it correctly suppresses the response for off-road placements.
\cref{fig:stitching_position} illustrates this qualitatively: trajectories visibly bend or shorten in response to road-placed sprites (Near, Far), while sky-placed sprites of the same size (Sky, SkyFar) leave the trajectory essentially unchanged.

\vspace{-0.5em}
\subsubsection{Ablation Study}
\label{sec:ablation_study_main}
\vspace{-0.5em}
The stitching results suggest that visual features shape our trajectory output. We now ask whether this behavior comes from next state prediction (NSP) or from the diffusion decoder alone.
We train every variant from scratch for four nuScenes epochs without the main model's additional language epoch, so absolute L2 values are not comparable to \cref{tab:main_exp}; all variants share the same diffusion IK head and change only what the decoder is conditioned on (\cref{tab:ablation_study_main}; full protocol in Appendix~\ref{sec:nsp_ik_disentanglement}).
Ego status and command alone already give a low L2 of $0.60/1.11$\,m but no response to the inserted obstacle (row 0+), so L2 alone cannot show visual grounding.
Current-state information (row i) gives a hazard--sky gap of only $5.8$ points. With NSP (row ii, ours), L2 improves to $0.53/0.98$\,m, the hazard rate rises from $71.3\%$ to $82.8\%$, the sky-placebo rate falls from $65.5\%$ to $45.2\%$, and the gap grows to $37.6$ points ($\sim 6.5\times$).
Feeding the ground-truth future (row iii) is not an upper bound: L2 rises to $2.58/4.05$\,m because the trajectory is shortened excessively.
More detailed ablation studies are provided in Appendix~\ref{sec:ablation_study}.

\vspace{-0.5em}
\section{Conclusion}
\label{sec:conclusion}
\vspace{-0.5em}

In this paper, we have addressed \emph{blind planning} of existing Driving VLAs.
We pointed out that, when trajectory generation is viewed as an inverse kinematics problem, existing VLAs lack a future visual state to anchor the trajectory, and we proposed a next visual state prediction objective that supplies this signal directly as supervision for the LLM, together with an Inverse Kinematics Network that decodes trajectories primarily from the current and future visual states, suppressing reliance on ego-status and text shortcuts.
Counterfactual interventions demonstrate improved hazard-specific visual responsiveness, while controlled ablations establish the contributions of the state-conditioning interface, next-state prediction, gradient isolation, and QA co-training. Our camera-only 0.5B model achieves state-of-the-art planning performance on NAVSIM-v1, NAVSIM-v2, and nuScenes. 
These results establish visual-state-conditioned action decoding as an effective design for grounding driving VLAs.

\begin{ack}
This work was supported by Institute of Information \& communications Technology Planning \& Evaluation (IITP) grant funded by the Korea government(MSIT) (No. RS-2024-00457882, National AI Research Lab Project), Basic Science Research Program through the National Research Foundation of Korea (NRF) funded by the MSIP (No. RS-2025-00520207), Institute of Information \& communications Technology Planning \& Evaluation (IITP) grant funded by the Korea government(MSIT) and Korea Evaluation Institute of Industrial Technology (KEIT) grant funded by the Korea government (MOTIE) (No.RS-2025-02217259, Development of self-evolving AI bias detection-correction-explain platform based on international multidisciplinary governance, Contribution Rate: 16.67\%), IITP grant funded by the Korea government(MSIT) and KEIT grant funded by the Korea government(MOTIE) (No. 2022-0-00680),  IITP grant funded by the Korea government(MSIT) and KEIT grant funded by the Korea government(MOTIE) (No. 2022-0-01045), next-generation CultureTech technology development (R\&D) project for cultural space AX transformation through the Korea Creative Content Agency grant funded by the Ministry of Culture, Sports and Tourism in 2026 (Project Name : Development of experiential library technology that allows users to speak out and experience with their whole body, Project Number : RS-2026-25508598, Contribution Rate : 16.67\%), and “Advanced GPU Utilization Support Program” funded by the Government of the Republic of Korea (Ministry of Science and ICT).

\end{ack}

\bibliographystyle{abbrvnat}
\bibliography{main}

\clearpage
\appendix
\section{Related Works}
\label{sec:related}

\subsection{End-to-end Planning}
End-to-end planning models take vehicle information such as the input image and ego status, and predict the trajectory along which the vehicle will proceed.
ST-P3~\cite{hu2022st} jointly optimizes perception/prediction/planning in vision-based end-to-end autonomous driving via BEV transformation and spatial-temporal feature learning.
VAD~\cite{jiang2023vad} represents the driving scene with a vectorized representation of agents and maps, simultaneously enhancing the efficiency and safety of planning.
UniAD~\cite{hu2023planning} integrates perception--prediction--planning into a single network and strengthens inter-module coordination through a planning-oriented design and unified queries.
InsightDrive~\cite{song2025insightdrive} organizes the attention required for planning through a language-guided scene representation that highlights ``regions important for driving'' in language.
Ego-MLP~\cite{zhai2023rethinking} analyzes how ego status can be a strong shortcut in open-loop planning and shows the possibility of high performance based solely on ego.
Building on this, Li \etal~\cite{li2024ego} show that a simple MLP using only physical state, without vision/point cloud, can be competitive on existing open-loop metrics, pointing out the loopholes of the evaluation metrics.
GenAD~\cite{zheng2024genad} presents a generative E2E paradigm that strengthens planning by predicting future scene/interaction changes from a ``generative modeling'' perspective.
DiffusionDrive~\cite{liao2025diffusiondrive, zou2025diffusiondrivev2} significantly reduces denoising steps with a truncated diffusion policy using multi-intent anchors and a cascade decoder, improving real-time multimodal trajectory planning performance on NAVSIM.
GoalFlow~\cite{xing2025goalflow} constrains the generative process with a goal point and combines scene-aware goal scoring with flow matching to efficiently produce high-quality multimodal trajectories with very few generation steps.

\subsection{Driving VLA}
Driving VLA models, like end-to-end planning models, take vehicle information such as input images and ego status, predict the trajectory along which the vehicle will proceed, and additionally perform scene description.
GPT-Driver~\cite{mao2023gpt} reformulates motion planning as language modeling and produces trajectories by generating coordinate tokens with a GPT-based model.
DriveGPT4~\cite{xu2024drivegpt4} proposes an interpretable end-to-end driving MLLM that jointly generates control signals and explanations through vision-language instruction tuning.
DriveLM~\cite{sima2024drivelm} presents data and benchmarks that elicit multi-step reasoning via graph-structured VQA, enhancing the generalization and interactivity of driving decision making.
DriveVLM~\cite{tian2024drivevlm} combines scene description, analysis, and hierarchical planning to strengthen reasoning-based planning in long-tail situations.
RDA-Driver~\cite{huang2024making} introduces a reasoning-decision alignment constraint to reduce the discrepancy between the chain-of-thought reasoning and the final decision, improving planning reliability.
DME-Driver~\cite{han2025dme} separates a VLM-based ``Decision Maker'' and a 3D-perception-based ``Executor'' to combine explainable decision making with stable control generation.
EMMA~\cite{hwang2024emma} processes camera input in language space with a multimodal LLM, directly mapping it to various driving outputs such as trajectories, objects, and road graphs.
OpenEMMA~\cite{xing2025openemma} is a lightweight open-source MLLM that takes ego-history and the front camera as input and predicts speed/curvature via chain-of-thought, integrating them into trajectory planning.
OpenDriveVLA~\cite{zhou2025opendrivevla} presents a VLA that combines 2D/3D instance-aware tokens, ego state, and language commands to produce spatially grounded driving actions/trajectories.
Impromptu VLA~\cite{chi2025impromptu} provides large-scale VLA data and open weights focused on corner cases, improving the robustness and open/closed-loop performance of VLAs.
GPVL~\cite{li2025generative} reduces the gap between perception and language through vision-language pretraining on BEV, and proposes a generative planning approach that autoregressively generates decisions and trajectories.
OmniDrive~\cite{wang2025omnidrive} builds a vision-language dataset that covers 3D driving understanding/planning based on counterfactual reasoning, strengthening VLM alignment and evaluation.
All such existing driving VLAs take an image as the initial condition and immediately perform trajectory prediction. We point out that this approach of existing driving VLAs, which lacks a terminal condition, is problematic from an inverse kinematics perspective, and we present a new architecture that addresses this problem.

\subsection{Driving World Models}
Driving World Models enable next state prediction for driving scenes, supporting tasks such as the simulation of driving scenarios.
LAW~\cite{li2024enhancing} predicts future latent features in a self-supervised manner conditioned on ego actions/trajectories, strengthening end-to-end driving without label dependence.
Drive-WM~\cite{wang2024driving} is a controllable multiview world model that generates future images for various maneuvers and supports safe planning with image-based rewards.
GAIA-1~\cite{hu2023gaia} presents a token-based generative world model that takes video, text, and action as input and generates controllable driving scenarios.
GAIA-2~\cite{russell2025gaia} proposes a controllable multi-view diffusion world model that simultaneously addresses multi-agent interactions and multi-camera consistency through structured conditioning.
Most recently, DriveVLA-W0~\cite{li2025drivevla} integrates future world prediction as an auxiliary objective for the VLA to enhance data scaling.

\paragraph{Distinction from DriveVLA-W0.}
Although both methods predict future states, our contribution changes the structure and causal role of future prediction beyond an input-restriction ablation.
In DriveVLA-W0~\cite{li2025drivevla}, the predicted future serves as a reasoning token within an autoregressive decoder, while trajectory generation remains autoregressive with direct access to ego status and text commands.
Our next visual state prediction (NSP), by contrast, supplies the boundary conditions of a separate action expert: the IK Network, a conditional diffusion decoder conditioned exclusively on the current and predicted next visual-state pair $(V_t, \hat V_{t+\Delta})$.
To our knowledge, this inverse-kinematics formulation, which anchors action decoding on a current/next visual-state pair, has not previously been explored in driving VLAs.
Mechanistically, future prediction becomes a required conditioning input at inference, whereas an auxiliary prediction objective alone leaves the trajectory decoder free to exploit the direct ego-status and command pathway.
The trajectory-loss analysis in \cref{eq:L0_decomp_ours,eq:ours_collapse_implies_const,eq:ours_bias_blowup} illustrates this distinction under its idealized assumptions: with direct access to $(E,C)$, the decoder can fit the bias term through these inputs alone, leaving visual collapse as a valid optimum; a decoder that ignores all variation in both visual-state inputs cannot fit a varying target mean. The controlled visual interventions in \cref{sec:stitching,sec:stitching_spatial} evaluate how the planner responds to changes in the visual scene.
The design is also effective at a compact model scale: DriveVLA-W0 uses a 7B backbone and reports 90.2 PDMS on NAVSIM-v1 and 86.5 EPDMS on NAVSIM-v2, while our 0.5B model trained with SFT$\rightarrow$GRPO achieves 92.2 and 90.6, respectively (\cref{tab:navsim_v1_exp,tab:navsim_v2_exp}).

Meanwhile, stand-alone generative world models such as GAIA-1/GAIA-2~\cite{hu2023gaia,russell2025gaia} are primarily aimed at photorealistic driving scene synthesis and simulation/data augmentation/counterfactual scenario generation, so planning is not a direct task and the target of next-state prediction is closer to pixel-level video synthesis.
In contrast, our next visual state prediction operates not in pixel space but in the visual token space on the LLM side, and does not aim to itself become a generative world model; its purpose is to supply the terminal condition the IK Network needs to recover trajectories.
In summary, while the GAIA family asks how the next scene may look, our predictor asks which visual state the ego should reach for the trajectory to be uniquely determined, and the two address different task formulations.

\section{Architecture Details}
\label{sec:appendix_arch}

\noindent\textbf{Visual encoder and LLM input.}
We extract patch-level visual tokens with DINOv3 ViT-S/16 \cite{simeoni2025dinov3} from the three front cameras (front, front-left, front-right), pass them through an MLP projector, and inject the resulting tokens into the LLM (Qwen2.5-0.5B-Instruct \cite{yang2024qwen25}).
On the input side, the visual tokens are placed between the special tokens \texttt{\textless{}scene\_start\textgreater{}} and \texttt{\textless{}scene\_end\textgreater{}}, and represented in text form by the placeholder \texttt{\textless{}SCENE\textgreater{}}.
The LLM additionally receives the ego status and text instruction encoded through an MLP, and (when present) a question about the driving scene wrapped between \texttt{\textless{}question\_start\textgreater{}} and \texttt{\textless{}question\_end\textgreater{}}.

\noindent\textbf{LLM output.}
The LLM is trained to emit the visual state of the future scene between \texttt{\textless{}scene\_start\textgreater{}} and \texttt{\textless{}scene\_end\textgreater{}} in its output, and the next visual state $\hat V_{t+\Delta}$ is read out from this region.
For QA, the answer appears between \texttt{\textless{}answer\_start\textgreater{}} and \texttt{\textless{}answer\_end\textgreater{}} and is supervised with next-token prediction.

\noindent\textbf{Inverse Kinematics Network.}
The IK Network is a cross-attention-based conditional diffusion model with hidden dimension $d_{\text{model}}=256$, $8$ attention heads, and $2$ cross-attention blocks.
The noisy trajectory serves as the query, and the current visual state tokens $V_t$ and the predicted next visual state tokens $\hat V_{t+\Delta}$ serve as keys/values; the timestep is added to the query through a sinusoidal embedding.
Diffusion training uses the cosine $\beta$-schedule \cite{nichol2021improved} and at inference we run $50$ DDPM \cite{ho2020denoising} sampling steps starting from $x_T \sim \mathcal{N}(0, I)$.

\section{Training Hyperparameters}
\label{sec:appendix_training}

\noindent\textbf{Optimization.}
We use AdamW \cite{loshchilov2019decoupled} with a cosine learning-rate schedule and linear warmup (warmup ratio $0.145$) and gradient clipping $1.0$ on both datasets.
Considering the difference in capacity and training stability between the LLM backbone and the IK Network, we apply different learning rates of $1\!\times\!10^{-5}$ and $1\!\times\!10^{-4}$, respectively.

\noindent\textbf{NAVSIM.}
We follow the 2-stage training procedure proposed by RecogDrive \cite{li2025recogdrive}: (i) supervised fine-tuning (SFT) on the navtrain split, followed by (ii) GRPO \cite{shao2024deepseekmath} reinforcement learning on the same split with PDMS-based reward signals.
The future prediction horizon is $4$\,s and the IK Network outputs $8$ waypoints.

\noindent\textbf{nuScenes.}
The main-table model is trained on the nuScenes train split for $4$ epochs and additionally on the language dataset for $1$ epoch.
The trajectory planning loss and the next visual state prediction loss are combined with weight $1.0$ each, and nuScenes-QA, nu-Caption, and nu-X are jointly trained as multi-task supervision through the same LLM.
The future prediction horizon is $3$\,s and the IK Network outputs $6$ waypoints.

\section{Input Prompt Format}
\label{sec:appendix_prompt}

We follow a Qwen-style ChatML conversation template with two roles, \texttt{system} and \texttt{user}, and condition the LLM on a structured prompt rather than free-form text.
This section makes the exact format explicit.

\noindent\textbf{Special tokens.}
On top of the base Qwen2.5 vocabulary we add the following \emph{additional special tokens}, each encoded as a single token:
\texttt{\textless{}scene\_start\textgreater{}}, \texttt{\textless{}scene\_end\textgreater{}}, \texttt{\textless{}SCENE\textgreater{}} (placeholder for the visual tokens $V_t$);
\texttt{\textless{}ego\_start\textgreater{}}, \texttt{\textless{}ego\_end\textgreater{}}, \texttt{\textless{}EGO\textgreater{}} (placeholder for the projected ego-status feature);
\texttt{\textless{}next\_start\textgreater{}}, \texttt{\textless{}next\_end\textgreater{}}, \texttt{\textless{}NEXT\_STATE\textgreater{}} (assistant-side placeholders into which $\hat V_{t+\Delta}$ is read out);
\texttt{\textless{}question\_start\textgreater{}}, \texttt{\textless{}question\_end\textgreater{}}, \texttt{\textless{}answer\_start\textgreater{}}, \texttt{\textless{}answer\_end\textgreater{}} for QA wrappers.
The visual placeholders \texttt{\textless{}SCENE\textgreater{}}, \texttt{\textless{}EGO\textgreater{}}, and \texttt{\textless{}VISUAL\_STATE\textgreater{}} are replaced by the corresponding feature embeddings at the input/output embedding stage rather than by tokenized text.

\noindent\textbf{System prompt.}
The system message specifies the role, the locations of the visual and ego placeholders, the coordinate convention, the two tasks (next visual state prediction and optional QA), and the strict output format:
\begin{quote}\small\ttfamily
You are DriveIK, a vision-language driving world model with latent actions.\\[2pt]
You are given:\\
- Current visual state tokens:\\
\textless scene\_start\textgreater\textless SCENE\textgreater\textless scene\_end\textgreater\\
- Ego status: \textless ego\_start\textgreater\textless EGO\textgreater\textless ego\_end\textgreater\\[2pt]
You will be given:\\
- Previous action trajectory.\\
- Optional nuScenes QA question.\\[2pt]
Coordinate convention (IMPORTANT):\\
- Ego-centric 2D BEV coordinates in meters; x is to the right, y is to the front.\\[2pt]
Your tasks:\\
1) Predict the NEXT visual state (one step ahead). To predict it, you MUST internally infer a feasible ego trajectory; DO NOT output the trajectory --- it is handled by a separate inverse-kinematics model.\\
2) If a QA question is provided, output the QA answer.\\[2pt]
Output format (STRICT). Output ONLY the blocks below.\\
Block 1 (next visual state): \textless visual\_start\textgreater\ \textless VISUAL\_STATE\textgreater\ ... \textless VISUAL\_STATE\textgreater\ \textless visual\_end\textgreater\\
Block 2 (optional QA): \textless answer\_start\textgreater SHORT\_ANSWER \textless answer\_end\textgreater\\
\end{quote}
The number of \texttt{\textless{}VISUAL\_STATE\textgreater{}} placeholders matches the spatial token count of $\hat V_{t+\Delta}$, and the LLM hidden states at those positions are read out as the predicted next visual state.

\noindent\textbf{User prompt.}
The user message follows an OpenDriveVLA-style structured layout that we build from the per-sample driving meta-data.
Concretely, each user prompt contains the lines below:
\begin{quote}\small\ttfamily
Ego status: \textless{}ego-status string\textgreater{}\\
Historical trajectory (last 2 seconds): [(x1,y1),(x2,y2),(x3,y3),(x4,y4)]\\
Mission goal: \{turn right $\mid$ turn left $\mid$ keep forward\}\\
Previous action: \textless{}previous action string\textgreater{}\\
\textless question\_start\textgreater \textless{}QA question\textgreater{} \textless question\_end\textgreater\quad(optional)
\end{quote}
The ego-status string is built from \texttt{gt\_ego\_lcf\_feat} and \texttt{gt\_ego\_his\_diff} fields of the nuScenes info dict and has the form
``\texttt{- Velocity (vx,vy): (\,$\cdot$\,,$\cdot$) - Heading Angular Velocity (v\_yaw): ($\cdot$) - Acceleration (ax,ay): ($\cdot$,$\cdot$) - Can Bus: ($\cdot$,$\cdot$) - Heading Speed: ($\cdot$) - Steering: ($\cdot$)}''.
The historical trajectory line is filled with $4$ recent ego positions in the same $(x_{\text{right}}, y_{\text{front}})$ convention used by the system prompt.
The mission goal is a one-of-three textual command derived from \texttt{gt\_ego\_fut\_cmd}.

\noindent\textbf{Assistant output.}
The assistant turn produces, in order, (i) the visual state block \texttt{\textless{}next\_start\textgreater{}}\,\texttt{\textless{}NEXT\_STATE\textgreater{}}\textsuperscript{$K$},\texttt{\textless{}next\_end\textgreater{}}, where $K$ matches the spatial token count of $\hat V_{t+\Delta}$, and (ii) optionally the answer block \texttt{\textless{}answer\_start\textgreater{}}\,\texttt{\textless{}SHORT\_ANSWER\textgreater{}},\texttt{\textless{}answer\_end\textgreater{}} when a QA question was provided.
Trajectories are \emph{not} produced by the LLM; the IK Network consumes the hidden states at the \texttt{\textless{}NEXT\_STATE\textgreater{}} positions, together with the current visual tokens, and outputs the trajectory through cross-attention diffusion.

\section{Detailed Analysis for \cref{sec:motivations}}
\label{sec:formulation}

This section derives the loss mechanism behind ego-status shortcut learning and explains how next visual state prediction and the IK Network address it.
The analysis identifies an ego-only optimum in the majority regime and the prediction error incurred by ignoring visual information under future-state supervision. Ego-status extrapolation calibrates the regime imbalance, while counterfactual stitching and controlled ablations establish the behavioral effects of our design.

\paragraph{The ego-predictable majority regime.}
For the short-horizon target trajectory $T_t$ ($3$--$4$\,s in our experiments), we assume
\begin{equation}
T_t \perp V_t \mid (E,C,\mathbb{I}_t=0),
\qquad
T_t=T_0(E,C)+\xi_t \quad\text{when }\mathbb{I}_t=0,
\label{eq:stylized_independence}
\end{equation}
with zero-mean noise independent of the inputs within this regime.
This assumption formalizes short-horizon cases in which the recorded trajectory is predictable from ego status and command, with little additional information supplied by vision.
Visual constraints already reflected in speed, yaw rate, and command make such trajectories readily extrapolated from $(E,C)$.
The vision-critical regime ($\mathbb{I}_t=1$) comprises cases where lane markings, traffic lights, speed limits, road geometry, or other visual evidence provide additional information that changes the target trajectory conditioned on $(E,C)$.
Thus, the indicator captures the incremental predictive role of vision across maneuver categories.
The exact decomposition below uses the conditional-independence idealization in \cref{eq:stylized_independence}.

\paragraph{Empirical calibration of the vision-critical minority.}
Vision-critical examples constitute only a small minority under our ego-status-gap proxy: approximately $10$--$30\%$, compared with a $70$--$90\%$ majority with smaller kinematic prediction errors.
Motivated by prior analyses of ego-status shortcuts~\cite{li2024ego,zhai2023rethinking}, \cref{sec:eps_sigma_estimates} quantifies this imbalance by extrapolating trajectories from current speed and yaw rate without visual input.
On nuScenes validation, the median ego-status gap is $1.286$\,m; only $30.0\%$ of samples exceed $2.14$\,m, and just $9.8\%$ exceed $5$\,m (\cref{tab:eps_thresholds}).
Thus, across these proxy thresholds, examples with large errors under ego-only extrapolation are consistently outnumbered by those with smaller errors.
This imbalance supplies far more training examples for learning an $(E,C)$ shortcut than for learning the visual corrections required by the vision-critical minority.
Counterfactual obstacle stitching further reveals weak visual responsiveness in three existing VLAs (\cref{tab:stitching_exp}), demonstrating the corresponding failure to adapt trajectories to changed visual conditions.
Together, the measured imbalance and observed blind planning support our central diagnosis: trajectory supervision is dominated in sample count by ego-predictable cases, leaving vision-critical learning concentrated in a small subset.
This motivates our two-part intervention: explicit future-state supervision and a decoder conditioned on visual states. The ablations below test their effects on trained models.

\subsection{Direct Regression Loss Decomposition}
\label{sec:formulation_direct}

We derive the bias--variance decomposition of direct trajectory regression stated in \cref{eq:var_penalty} of \cref{sec:motivations}. Under the majority-regime assumption, this decomposition shows that ignoring visual features is an optimal solution to the regime-specific loss, exposing the shortcut admitted by direct regression.

\paragraph{Notation.}
Let $V_t$ denote the output of a frozen visual encoder, $E$ the ego status, $C$ the text command, and $\mathbb{I}\in\{0,1\}$ the vision-critical indicator.
We continue to use the target trajectory decomposition $T_t = T_0(E,C) + \mathbb{I}\cdot\Delta(V,E,C) + \xi$ (\cref{eq:traj_decompose}) and the predictor decomposition
\begin{equation}
\bar f_\theta(E,C) \triangleq \mathbb{E}\!\left[f_\theta(V,E,C)\mid E,C,\mathbb{I}{=}0\right],
\qquad
\tilde f_\theta(V,E,C) \triangleq f_\theta(V,E,C) - \bar f_\theta(E,C).
\label{eq:f_decomp_app}
\end{equation}
Since $\bar f_\theta(E,C)$ is by definition the conditional expectation of $f_\theta$ given $(E,C,\mathbb{I}{=}0)$, it is constant given $(E,C)$, 
and by linearity of conditional expectation $\mathbb{E}[\tilde f_\theta\mid E,C,\mathbb{I}{=}0] = \mathbb{E}[f_\theta\mid E,C,\mathbb{I}{=}0] - \bar f_\theta(E,C) = 0$, i.e., $\tilde f_\theta$ is the conditional residual relative to that mean.
We additionally assume that the noise $\xi$ is zero-mean and (conditionally) independent of $(E,C,V)$ and $\tilde f_\theta$.

\paragraph{Step-by-step derivation of the bias--variance decomposition.}
Under $\mathbb{I}=0$, the target is $T_0(E,C)+\xi$, so we expand the conditional MSE as follows:
\begin{align}
\mathcal{L}_0(\theta)
&= \mathbb{E}\!\left[\|T_0(E,C)+\xi - f_\theta(V,E,C)\|_2^2 \,\big|\, E,C,\mathbb{I}{=}0\right]
\label{eq:app_L0_start}\\
&= \mathbb{E}\!\left[\|\bigl(T_0(E,C)-\bar f_\theta(E,C)\bigr) - \tilde f_\theta(V,E,C) + \xi\|_2^2 \,\big|\, E,C,\mathbb{I}{=}0\right]
\label{eq:app_L0_subst}\\
&= \|T_0(E,C)-\bar f_\theta(E,C)\|_2^2
   + \mathbb{E}\!\left[\|\tilde f_\theta\|_2^2 \,\big|\, E,C,\mathbb{I}{=}0\right]
   + \mathbb{E}\!\left[\|\xi\|_2^2\right].
\label{eq:app_L0_final}
\end{align}
\eqref{eq:app_L0_subst} is obtained by simply substituting $f_\theta=\bar f_\theta+\tilde f_\theta$ (\cref{eq:f_decomp_app}).
In the transition to \eqref{eq:app_L0_final}, the three cross terms that arise are all expanded as inner products and then cancelled as follows:
\begin{itemize}
\item $-2\,\mathbb{E}\!\left[\langle T_0(E,C)-\bar f_\theta(E,C),\, \tilde f_\theta\rangle \,\big|\, E,C,\mathbb{I}{=}0\right]=0$:
because $(T_0-\bar f_\theta)$ is constant given $(E,C)$ and $\mathbb{E}[\tilde f_\theta\mid E,C,\mathbb{I}{=}0]=0$.
\item $\;\;2\,\mathbb{E}\!\left[\langle T_0(E,C)-\bar f_\theta(E,C),\, \xi\rangle \,\big|\, E,C,\mathbb{I}{=}0\right]=0$:
for the same reason, $(T_0-\bar f_\theta)$ is constant and $\mathbb{E}[\xi\mid E,C,\mathbb{I}{=}0]=0$.
\item $-2\,\mathbb{E}\!\left[\langle \tilde f_\theta,\, \xi\rangle \,\big|\, E,C,\mathbb{I}{=}0\right]=0$:
by the assumption that $\xi$ is zero-mean and (conditionally) independent of $\tilde f_\theta$.
\end{itemize}
This treatment of cross terms is the same as that used in the next-state predictor analysis (\cref{eq:s0_substitute}--\cref{eq:s0_final}) and the IK Network analysis (\cref{eq:L0_decomp_ours}) of this section.

\subsection{Next Visual State Prediction}
\label{sec:formulation_state}

In this section, we mathematically analyze how the next visual state prediction objective introduced in \cref{sec:motivations} penalizes visual collapse.
Through a bias--variance decomposition of the next visual state prediction loss $\mathcal{L}_{\mathrm{state}}$, we show that ignoring visual features leaves scene-dependent prediction error that an ego-only predictor cannot reduce.
When the model can predict this residual variation from current visual features, visual collapse is \emph{strictly suboptimal}: reducing the loss further requires using visual information beyond $(E,C)$.

\paragraph{Notation.}
For brevity, let $V_t \triangleq \mathrm{enc}(V_t)$ denote the output of a frozen visual encoder, $E$ the ego status, $C$ the text command, and $V_{t+\Delta}$ the visual state target token $\Delta$ seconds in the future.
We denote the LLM's next-state predictor as $w_\phi(V_t, E, C)$ and retain the vision-critical indicator $\mathbb{I}\in\{0,1\}$ defined above. This indicator is not an input mask; the visual stream is available in both regimes.

\paragraph{Bias--variance decomposition.}
We define the conditional means and residuals of the predictor and target as follows:
\begin{equation}
\bar w_\phi(E,C) \triangleq \mathbb{E}\bigl[w_\phi(V_t,E,C)\mid E,C,\mathbb{I}{=}0\bigr],
\qquad
\tilde w_\phi \triangleq w_\phi(V_t,E,C) - \bar w_\phi(E,C),
\label{eq:w_decomp}
\end{equation}
\begin{equation}
\bar V(E,C) \triangleq \mathbb{E}\bigl[V_{t+\Delta}\mid E,C,\mathbb{I}{=}0\bigr],
\qquad
\tilde V \triangleq V_{t+\Delta} - \bar V(E,C).
\label{eq:V_decomp}
\end{equation}
By definition, $\mathbb{E}[\tilde w_\phi\mid E,C,\mathbb{I}{=}0]=0$ and $\mathbb{E}[\tilde V\mid E,C,\mathbb{I}{=}0]=0$.

Using these, the next-state loss $\mathcal{L}_{\mathrm{state},0}$ (the loss defined under $\mathbb{I}=0$) can be expanded as:
\begin{align}
\mathcal{L}_{\mathrm{state},0}
&= \mathbb{E}\!\left[\|V_{t+\Delta}-w_\phi(V_t,E,C)\|_2^2 \,\big|\, E,C,\mathbb{I}{=}0\right]
\label{eq:s0_start}\\
&= \mathbb{E}\!\left[\|(\bar V-\bar w_\phi) + (\tilde V-\tilde w_\phi)\|_2^2 \,\big|\, E,C,\mathbb{I}{=}0\right]
\label{eq:s0_substitute}\\
&= \|\bar V(E,C)-\bar w_\phi(E,C)\|_2^2
   + \mathbb{E}\!\left[\|\tilde V-\tilde w_\phi\|_2^2 \,\big|\, E,C,\mathbb{I}{=}0\right].
\label{eq:s0_final}
\end{align}
In \eqref{eq:s0_substitute} we have simply substituted $V_{t+\Delta}=\bar V+\tilde V$ and $w_\phi=\bar w_\phi+\tilde w_\phi$.
In the transition to \eqref{eq:s0_final}, the cross term cancels because $(\bar V-\bar w_\phi)$ is constant given $(E,C)$ and, by construction, both $\tilde V$ and $\tilde w_\phi$ are residuals relative to their conditional means: $\mathbb{E}[\tilde V \mid E,C,\mathbb{I}{=}0]=0$ and $\mathbb{E}[\tilde w_\phi \mid E,C,\mathbb{I}{=}0]=0$. Hence by linearity of conditional expectation, $\mathbb{E}[\tilde V - \tilde w_\phi \mid E,C,\mathbb{I}{=}0]=0$.
That is, the loss neatly splits into a \emph{bias term} $\|\bar V - \bar w_\phi\|_2^2$ and a \emph{conditional variance term} $\mathbb{E}[\|\tilde V - \tilde w_\phi\|_2^2]$.

\paragraph{Lower bound under visual collapse.}
We now define ``visual collapse'' as the state in which the LLM does not effectively make use of $V_t$, i.e., the state in which the output of $w_\phi$ is invariant with respect to $V_t$:
\begin{equation}
w_\phi(V_t,E,C) \approx g_\phi(E,C) \;\;\forall V_t
\quad\Longleftrightarrow\quad
\tilde w_\phi \approx 0.
\label{eq:collapse_def}
\end{equation}
We also reflect the fact that, even under the same $(E,C)$, the future visual state varies due to the inherent diversity of scenes, by defining the data-intrinsic visual diversity as
\begin{equation}
\sigma_V^2 \triangleq \mathbb{E}_{E,C}\!\bigl[\mathrm{Var}(V_{t+\Delta}\mid E,C)\bigr] > 0.
\label{eq:sigma_V}
\end{equation}

Under these two definitions, when collapse occurs ($\tilde w_\phi=0$) the bias--variance decomposition simplifies as follows:
\begin{align}
\mathcal{L}_{\mathrm{state},0}
&= \|\bar V(E,C)-\bar w_\phi(E,C)\|_2^2 + \mathbb{E}\!\left[\|\tilde V - \tilde w_\phi\|_2^2 \,\big|\, E,C,\mathbb{I}{=}0\right]
\label{eq:state0_decomp}\\
&\geq \mathbb{E}\!\left[\|\tilde V - \tilde w_\phi\|_2^2 \,\big|\, E,C,\mathbb{I}{=}0\right]
\label{eq:state0_drop_bias}\\
&= \mathbb{E}\!\left[\|\tilde V\|_2^2 \,\big|\, E,C,\mathbb{I}{=}0\right]
\label{eq:state0_collapse}\\
&= \mathrm{Var}(V_{t+\Delta}\mid E,C,\mathbb{I}{=}0).
\label{eq:state0_var}
\end{align}
\eqref{eq:state0_drop_bias} simply drops the bias term, which is non-negative; \eqref{eq:state0_collapse} is the result of substituting $\tilde w_\phi=0$ from the collapse definition \eqref{eq:collapse_def};
finally, \eqref{eq:state0_var} is a direct consequence of $\bar V(E,C)$ being defined as a conditional mean, expressing the conditional variance.

The last equality bounds the conditional loss of a collapsed predictor; averaging it over $(E,C)$ within the same regime yields the corresponding regime-specific variance bound.
When the predictor can recover a nonzero component of this residual variation from $V_t$, using vision reduces the attainable loss below the best ego-only prediction, making visual collapse strictly suboptimal.
This is the central difference from direct trajectory regression: NSP rewards predictive visual information even within the ego-predictable trajectory regime.
A loss below the bound, evaluated with the same conditioning and feature representation, rules out visual collapse.

\subsection{Inverse Kinematics Formulation}

The IK Network connects future-state prediction to trajectory generation by decoding actions from the current and predicted visual states.
Existing VLAs admit a direct ego-status shortcut in trajectory regression (\cref{sec:motivations}); our decoder removes this bypass and receives task context through the supervised state interface.
Together with NSP supervision and trajectory-gradient isolation, this interface makes state prediction the representation on which action decoding operates.

\paragraph{Decomposition of the trajectory loss.}
Our IK Network $h_\psi$ decodes the trajectory $T$ taking only the current visual state $V_t$ and the next visual state $\hat V_{t+\Delta}$ predicted by the LLM as input.
Letting $T_0(E,C)$ denote the mean of the ``ideal'' trajectory determined from $(E,C)$, and $\xi$ the zero-mean noise inherent in the trajectory, the trajectory loss under $\mathbb{I}=0$ is decomposed as follows:
\begin{align}
\mathcal{L}_{\mathrm{traj},0}(\psi)
&= \mathbb{E}\!\left[\|T_0(E,C) + \xi - h_\psi(V_t, \hat V_{t+\Delta})\|_2^2 \,\big|\, E,C,\mathbb{I}{=}0\right] \nonumber \\
&= \underbrace{\|T_0(E,C) - \bar h_\psi(E,C)\|_2^2}_{\text{bias}}
   + \underbrace{\mathbb{E}\!\left[\|\tilde h_\psi\|_2^2 \,\big|\, E,C,\mathbb{I}{=}0\right]}_{\text{variance}}
   + \underbrace{\mathbb{E}[\|\xi\|_2^2]}_{\text{irreducible noise}}.
\label{eq:L0_decomp_ours}
\end{align}
Here, $\bar h_\psi(E,C)$ and $\tilde h_\psi$ are the conditional mean and residual defined in the same manner as for $w_\phi$.

While the \emph{form} of the equation is identical to that of the decomposition obtained for the existing structure, the \emph{coupling} between the two terms is decisively different.

\paragraph{Decisive difference between the existing and the proposed structures.}
In existing VLAs, since the trajectory predictor $f_\theta$ takes $(V_t, E, C)$ all directly as input, the conditional mean $\bar f_\theta(E,C)$ can be computed using only the $(E,C)$ token pathway.
Hence, even if the model chooses $\tilde f_\theta = 0$ (ignoring visual features), the learning of $\bar f_\theta$ is unaffected, and the two terms can be minimized \emph{independently}, making it a valid optimal solution to ignore visual features.

In contrast, our IK Network $h_\psi$ does not take $(E, C)$ directly as input and only takes $(V_t, \hat V_{t+\Delta})$.
The decoder therefore receives task information through the current and predicted visual-state tokens.
The predicted state $\hat V_{t+\Delta}=w_\phi(V_t,E,C)$ integrates scene information with the ego and command context needed for planning.
The stop-gradient interface in \cref{sec:ik_gradient_isolation} prevents the trajectory loss from adapting these tokens, while NSP supervises their agreement with future visual features.

Fully ignoring both conditioning inputs forces the decoder to output a single constant $c$:
\begin{equation}
h_\psi(V_t,\hat V_{t+\Delta})=c\quad\text{for all conditioning pairs}
\;\Longrightarrow\;
\bar h_\psi(E,C)=c.
\label{eq:ours_collapse_implies_const}
\end{equation}
Then the best attainable average bias is
\begin{equation}
\min_c\,\mathbb{E}\!\left[\|T_0(E,C)-c\|_2^2\mid\mathbb{I}=0\right]
=\mathrm{Var}\!\left(T_0(E,C)\mid\mathbb{I}=0\right)>0,
\label{eq:ours_bias_blowup}
\end{equation}
provided the target conditional mean varies across $(E,C)$ in this regime.

These equations show that an IK Network that ignores both state inputs must predict the same trajectory for every driving condition. It therefore incurs a nonzero average bias when the target mean trajectory varies across $(E,C)$.
This is stronger than $\tilde h_\psi=0$: zero residual at fixed $(E,C)$ still allows the mean prediction to vary across driving conditions.
Our method uses NSP to supervise the predicted visual states and stop-gradient to prevent the trajectory loss from updating the state representations. We evaluate visual responsiveness with stitching and the safety benefit of gradient isolation with a controlled ablation (\cref{tab:stitching_exp,tab:ik_gradient_flow}).



\subsection{Empirical Estimates of $\varepsilon$ and $\sigma_V^2$ on nuScenes}
\label{sec:eps_sigma_estimates}

The arguments in \cref{sec:formulation_direct,sec:formulation_state} hinge on two scalar quantities that we have so far treated only symbolically.
The first is the fraction $\varepsilon \triangleq \mathbb{P}(\mathbb{I}_t = 1)$ of vision-critical samples in the trajectory decomposition $T_t = T_0(E,C) + \mathbb{I}_t \cdot \Delta(V,E,C) + \xi$ (\cref{eq:traj_decompose}), which sets the weight of the only term in the mixture loss (\cref{eq:loss_mixture}) that pushes the trajectory predictor to depend on $V$.
The second is the data-intrinsic visual diversity $\sigma_V^2 = \mathbb{E}_{E,C}[\mathrm{Var}(V_{t+\Delta}\mid E,C)] > 0$ from \cref{eq:sigma_V}, the lower bound below which the next-state loss cannot drop without using $V_t$ (\cref{eq:state0_var}).
This section reports empirical proxies for these quantities on the nuScenes validation split using the DINOv3 ViT-S/16 features adopted in our experiments (front-left, front, front-right; $540$ tokens of dimension $384$ per frame; $\Delta = 3$\,s, equivalently $6$ keyframes ahead).

\paragraph{Estimating $\varepsilon = \mathbb{P}(\mathbb{I}_t = 1)$.}
We use the following operational definition.
For each validation sample, we read the ego state at the current frame from the nuScenes info dict field \texttt{gt\_ego\_lcf\_feat}, a per-frame vector of ego-vehicle states (the same field that supplies the ego-status string in the input prompt).
We use only three of its entries: the longitudinal velocity $v_x$, the lateral velocity $v_y$, and the yaw rate $v_\text{yaw}$.
Starting from the current pose, we roll these three values forward over the $3$-second horizon with a constant-velocity, constant-yaw-rate motion model, which gives a kinematic trajectory that uses no visual information.
The \emph{ego-status gap} is the ADE between this kinematic trajectory and the GT trajectory over the same $3$\,s horizon.
A small gap means that ego status alone nearly determines the future trajectory; a large gap means that it does not.
We operationalize the vision-critical subset using an ego-status gap of at least $\tau$, yielding a measurable proxy for the latent regime indicator in \cref{eq:stylized_independence}.
The marginal distribution of the gap on the nuScenes val split ($n = 6{,}019$) has mean $2.230$\,m, median $1.286$\,m, $70$th percentile $2.139$\,m, and $90$th percentile $4.888$\,m.
\Cref{tab:eps_thresholds} reports $\hat{\varepsilon}(\tau) = \tfrac{1}{n}\sum_{i} \mathbf{1}[\text{gap}_i \geq \tau]$ for several thresholds.

\begin{table}[htbp]
\centering
\small
\caption{Empirical $\hat{\varepsilon}$ on the nuScenes val split ($n = 6{,}019$) at several gap thresholds. The threshold $\tau = 2.14$\,m selects the top $30\%$ of samples by ego-status gap.}
\label{tab:eps_thresholds}
\begin{adjustbox}{max width=\linewidth}
\begin{tabular}{c|ccccccc}
\toprule
$\tau$ (m) & $0.50$ & $1.00$ & $1.50$ & $2.00$ & $\mathbf{2.14}$ & $3.00$ & $5.00$ \\
\midrule
$\hat{\varepsilon}$ & $0.763$ & $0.584$ & $0.442$ & $0.328$ & $\mathbf{0.300}$ & $0.187$ & $0.098$ \\
\bottomrule
\end{tabular}
\end{adjustbox}
\end{table}

At $\tau=2.14$\,m the proxy fraction is $0.300$, and at $\tau=5$\,m it is $0.098$.
Across these thresholds, the proxy-defined vision-critical subset accounts for only $10$--$30\%$ of samples, leaving a $70$--$90\%$ majority with smaller kinematic errors.
This sample imbalance concentrates the trajectory-supervision incentive for visual corrections in a minority of training examples.
NSP addresses this imbalance by supervising scene-dependent future visual content across both trajectory regimes.

\paragraph{Estimating $\sigma_V^2 = \mathbb{E}_{E,C}[\mathrm{Var}(V_{t+\Delta}\mid E,C)]$.}
Here we measure how much the scene $3$ seconds later still varies when the route command is known.
For each val sample $t$, we follow the \texttt{next} links in the cached info dict to find the sample $\Delta = 6$ keyframes ahead ($3$\,s later).
We take its DINOv3 feature $V_{t+\Delta} \in \mathbb{R}^{540 \times 384}$ and compute the mean and variance of every feature entry with Welford's online algorithm.
Of the $6{,}019$ val samples, $5{,}119$ have cached features at both the current time and $+3$\,s, so we use only these.
Conditioning on $(E, C)$ exactly is impractical because both are continuous and high-dimensional.
Instead, we use the three-way route command \texttt{gt\_ego\_fut\_cmd} (right / forward / left) in place of $C$.
We report two numbers: the overall variance $\mathrm{Var}(V_{t+\Delta})$, and the average variance within each command, $\mathbb{E}_C[\mathrm{Var}(V_{t+\Delta}\mid C)]$.

\begin{table}[htbp]
\centering
\small
\caption{Variance of the future DINOv3 feature $V_{t+\Delta}$ on the nuScenes val split ($n=5{,}119$). Each value is the sum of the variances of all $540 \times 384 = 207{,}360$ feature entries. We condition only on the route command, not on the full conditioning in \cref{eq:state0_var}, so these values are approximate proxies.}
\label{tab:sigma_v}
\begin{adjustbox}{max width=\linewidth}
\begin{tabular}{l|c}
\toprule
quantity & sum of variances \\
\midrule
Overall variance $\mathrm{Var}(V_{t+\Delta})$                                         & $1.162\!\times\!10^{4}$ \\
Variance within a command $\mathbb{E}_C[\mathrm{Var}(V_{t+\Delta}\mid C)]$ ($\sigma_V^2$ proxy) & $1.158\!\times\!10^{4}$ \\
Variance between commands                                                              & $34$ \\
\bottomrule
\end{tabular}
\end{adjustbox}
\end{table}

The within-command variance is $0.997$ times the overall variance.
In other words, the route command explains only about $0.3\%$ of the visual variation.
The within-command variance is about $1.16\times10^4$ in total, or $5.6\times10^{-2}$ per feature entry.
It shows how much the future scene still varies once the command is known.
The bound in \cref{eq:state0_var} also conditions on ego status and the regime, so the two are not exactly the same quantity.


\paragraph{Joint reading.}
The ego-status gap distribution establishes a pronounced sample imbalance under the proxy, while the visual-feature statistics reveal substantial scene variation beyond the route command.
These measurements explain the motivation for supervising future visual states: trajectory targets are often readily extrapolated from ego status, whereas visual targets retain scene-dependent content.
Counterfactual stitching and controlled ablations then establish the contribution in model behavior: stronger hazard-specific responses and improved planning with the proposed state interface.

\section{Gradient Isolation at the IK Conditioning Interface}
\label{sec:ik_gradient_isolation}

\paragraph{Stop-gradient conditioning.}
Restricting the IK Network's forward inputs to visual-state tokens alone does not prevent trajectory-loss gradients from modifying the predicted state during joint optimization.
Our implementation therefore applies stop-gradient to both conditioning inputs, the current visual state $V_t$ and the predicted next visual state $\hat V_{t+\Delta}$.
Writing $\operatorname{sg}(\cdot)$ for an operation that preserves its input in the forward pass and has zero derivative, the diffusion decoder receives
\begin{equation}
\epsilon_\psi\!\left(x_s,s,
\operatorname{sg}(V_t),
\operatorname{sg}(\hat V_{t+\Delta})\right),
\qquad
\nabla_\phi \mathcal{L}_{ik}=0,
\label{eq:ik_stop_gradient}
\end{equation}
where $s$ denotes the diffusion timestep and $\phi$ denotes the LLM parameters.
Thus, the IK trajectory loss updates the diffusion decoder but is never propagated to the LLM through its conditioning inputs.
The predicted next visual state is directly supervised by the NSP MSE objective against ground-truth future DINOv3 features in the projected visual-token space, rather than by the trajectory loss.
Auxiliary QA supervision can still update the shared LLM parameters, as described in \cref{sec:appendix_training}.
This separation closes the trajectory-gradient channel through which the diffusion head could otherwise reshape the predicted state to encode a planning shortcut.
It provides the same isolation from trajectory-loss updates as training the state predictor first and then training the action decoder with the predictor frozen; it does not require the two modules to be optimized in separate chronological stages.

\paragraph{Controlled gradient-flow ablation.}
To measure the effect of this design, we compare two models on the NAVSIM \texttt{navtest} split with identical data, training schedules, and architectures, changing only whether IK trajectory gradients are blocked at the conditioning interface or allowed to flow back to the LLM.
\Cref{tab:ik_gradient_flow} reports the results.

\begin{table}[htbp]
\centering
\small
\caption{Effect of IK trajectory-gradient flow to the LLM on  \texttt{NAVSIM v1} and \texttt{v2}. The two variants differ only in whether gradients are blocked at the IK conditioning inputs. Higher is better for all metrics.}
\label{tab:ik_gradient_flow}
\begin{adjustbox}{max width=\linewidth}
\begin{tabular}{lcccccc}
\toprule
IK trajectory gradient to LLM & PDMS v1 $\uparrow$ & EPDMS v2 $\uparrow$ & NC $\uparrow$ & DAC $\uparrow$ & TTC $\uparrow$ & EP $\uparrow$ \\
\midrule
Detached (ours) & 90.29 & 89.25 & 99.18 & 96.22 & 99.00 & 87.60 \\
Backflow allowed & 86.24 & 84.91 & 97.11 & 93.89 & 96.65 & 88.11 \\
\bottomrule
\end{tabular}
\end{adjustbox}
\end{table}

Allowing gradient backflow reduces PDMS by $4.05$ points and EPDMS by $4.34$ points.
The reported safety metrics all deteriorate: NC decreases by $2.07$ points, DAC by $2.33$ points, and TTC by $2.35$ points, while ego progress remains similar, increasing slightly by $0.51$ points.
With data, architecture, and training schedule fixed, blocking trajectory gradients improves all three safety metrics and both aggregate planning scores while preserving comparable progress. This ablation establishes gradient isolation as an effective component of the state interface.

\section{Note on the OpenDriveVLA* Score in NAVSIM Benchmarks}
\label{sec:opendrivevla_score_range}
OpenDriveVLA* records PDMS/EPDMS of $73.2/70.2$ in \cref{tab:navsim_v1_exp,tab:navsim_v2_exp}.
It uses the same 0.5B Qwen2.5 backbone and DINOv3 ViT-S/16 encoder as our model, with matched data and training schedules, and preserves direct trajectory text output.
This matched comparison isolates the combined NSP and IK design from differences in backbone scale, visual encoder, data, and training schedule: our model improves by $19.0$ PDMS and $20.4$ EPDMS.
The broader benchmark tables additionally compare against VLAs with their reported backbone, encoder, and decoding configurations.
OpenDriveVLA* is the matched NAVSIM implementation; the original nuScenes model uses a UniAD BEV encoder \cite{hu2023planning}. The benchmark tables identify these distinct configurations.

\section{GradCAM Visual Grounding Analysis and Methodology}
\label{sec:gradcam_methodology}

This section summarizes the detailed measurement protocol for the visual grounding analysis in \cref{sec:gradcam_overlap}.
We adapt the backpropagation target and hook layer to each model's trajectory output form and report common object-overlap metrics.
The overlap metrics quantify alignment between model sensitivity and scene objects, complementing the direct behavioral evidence from obstacle stitching and position--size controls in \cref{sec:stitching,sec:stitching_spatial}.
The backpropagation protocol below specifies the architecture-dependent meaning of each saliency target.

\subsection{Visual Grounding Results}
\label{sec:gradcam_overlap}
\vspace{-0.5em}
The stitching experiment in \cref{sec:stitching} shows that visual perturbations \emph{change} the trajectory; here we examine where gradient-based saliency is concentrated.
On the nuScenes validation set, we measure how much each model's GradCAM \cite{selvaraju2020grad} overlaps with on-road objects (vehicles and pedestrians).
For each sample, we project GT 3D bounding boxes from the front camera onto the 2D image plane to obtain an object mask $M$, and obtain a per-sample saliency map $S$ using the architecture-specific backpropagation targets described in \cref{sec:gradcam_backprop_target}.
We report three complementary metrics: (1) \emph{Pointing Mass} (PM), the fraction of CAM energy that falls inside $M$; (2) \emph{Average Precision} (AP), a threshold-free ranking score treating $M$ as positives and $S$ as scores; and (3) \emph{IoU/F1 @ top-20\%}, region-level overlap on the top-20\% saliency pixels.
The following subsections detail the measurement protocol. We compare against OpenDriveVLA \cite{zhou2025opendrivevla}, Alpamayo-R1 \cite{wang2025alpamayo}, and Impromptu-VLA \cite{chi2025impromptu}.

\begin{table}[htbp]
\caption{
Visual grounding via GradCAM--object overlap on nuScenes \emph{val set}.
}

\centering
\resizebox{\linewidth}{!}{%
\begin{tabular}{lcccc}
\toprule
\textbf{Model} & \textbf{Pointing Mass}$\uparrow$ & \textbf{Avg. Precision}$\uparrow$ & \textbf{IoU@top20}$\uparrow$ & \textbf{F1@top20}$\uparrow$  \\
\midrule
Impromptu-VLA 3B \cite{chi2025impromptu} & 0.103 & 0.109 & 0.067 & 0.121  \\
Alpamayo-R1 10B \cite{wang2025alpamayo}  & 0.178 & 0.202 & 0.093 & 0.161  \\
OpenDriveVLA 0.5B \cite{zhou2025opendrivevla} & \underline{0.205} & \underline{0.243} & \underline{0.105} & \underline{0.177}  \\
\midrule
Ours & \textbf{0.268} & \textbf{0.281} & \textbf{0.145} & \textbf{0.236}  \\
\bottomrule
\end{tabular}%
}
\vspace{2mm}
\label{tab:gradcam_overlap}
\end{table}
\begin{figure*}[t]
\centering
   \includegraphics[width=1.0\linewidth,height=0.72\textheight,keepaspectratio]{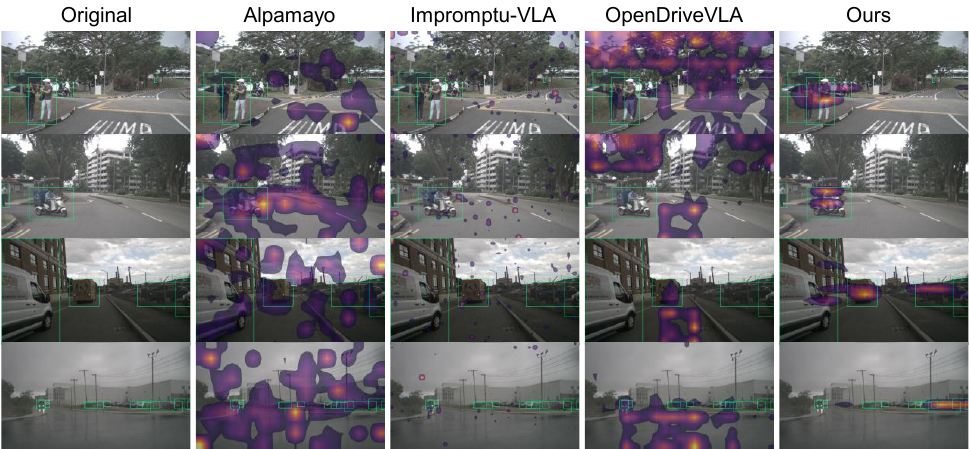}
\caption{
Qualitative comparison of GradCAM saliency on nuScenes \emph{val set}.
Each row corresponds to a single driving scene, and each column shows the GradCAM of Alpamayo-R1 \cite{wang2025alpamayo}, Impromptu-VLA \cite{chi2025impromptu}, OpenDriveVLA \cite{zhou2025opendrivevla}, and our model overlaid on the input image for that scene.
}
\label{fig:gradcam_overlap}
\end{figure*}

\paragraph{Results.}
As shown in \cref{tab:gradcam_overlap}, our method achieves the highest object alignment on all three metrics. In Pointing Mass we record $0.268$, $+31\%$ over the second-place OpenDriveVLA ($0.205$) and up to $+160\%$ over Impromptu-VLA ($0.103$); on AP we obtain $0.281$ ($+16\%$ over OpenDriveVLA), and the 95\% bootstrap confidence intervals do not overlap.
\cref{fig:gradcam_overlap} confirms this qualitatively: existing models disperse GradCAM over road surface, sky, and background, while ours concentrates it on planning-relevant objects such as front vehicles and pedestrians, consistent with the analysis in \cref{sec:motivations}.

\subsection{Metric Definitions}
\label{sec:gradcam_metric_def}

Given a per-sample saliency map $S \in \mathbb{R}_{\geq 0}^{H\times W}$ and an object mask $M \in \{0,1\}^{H\times W}$ obtained by projecting GT 3D bounding boxes (categories \texttt{vehicle.*} and \texttt{human.pedestrian.*}, with boxes containing fewer than one LiDAR point excluded) onto the 2D image plane, we report three complementary metrics.

\textbf{(1) Pointing Mass} is a \emph{threshold-free} metric measuring how much of the model's ``attention budget'' is allocated to actual object regions, defined as the ratio of CAM energy located inside the object mask to the total CAM energy:
\[
\text{PM} = \frac{\sum_{(i,j)\in M} S_{ij}}{\sum_{(i,j)} S_{ij}}.
\]
This metric measures only ``\emph{where the model is looking}'' regardless of the absolute scale of the saliency map.

\textbf{(2) Average Precision (AP)} is the precision-recall AUC averaged over all thresholds, treating the object mask as positive labels and $S$ as the ranking score.
Since no threshold is artificially fixed, it answers the question ``are pixels with higher saliency values more likely to actually lie on objects?''

\textbf{(3) IoU \& F1 @ top-20\%} treats the top-20\% pixels of the saliency map (i.e., pixels above the per-image $80$-percentile $p_{80}$) as binary positives and measures spatial overlap with the object mask in terms of IoU and F1, complementing the ``ranking consistency'' captured by PM and AP with a region-level agreement.

The three metrics are based on different assumptions (energy ratio, threshold-free ranking, hard mask overlap), so a consistent advantage across all metrics indicates an essential improvement in grounding quality rather than single-metric hacking.

\noindent\textbf{Scale invariance.}
The raw gradient magnitudes across different architectures cannot be directly compared, so all three metrics above are constructed to be robust to cross-architecture gradient scale differences:
(i) PM is a ratio of sums and is therefore invariant to multiplication by a positive constant;
(ii) AP is a rank-based score and is invariant to any positive monotonic transformation of $S$;
(iii) IoU/F1 @ top-20\% uses the per-image $80$-percentile as the threshold, so the binarization is unchanged under multiplication by a positive constant.
In addition, all CAMs are min-max normalized per camera per image to the range $[0,1]$ before being used in metric computation.

\subsection{Backpropagation Target}
\label{sec:gradcam_backprop_target}

GradCAM uses the gradient of some scalar $\mathcal{S}$ with respect to the visual token input as visual saliency, so the choice of $\mathcal{S}$ determines the meaning of the measurement.
Where available, we use the model's native trajectory supervision objective as the backpropagation scalar; for Alpamayo-R1, we use the proxy described below.
Each scalar measures sensitivity in its model's computation; comparisons use the normalized spatial-overlap metrics above, with shared input interventions providing the common behavioral evaluation.

\noindent\textbf{Ours (CrossAttentionDiffusionIK).}
Instead of an ELBO or marginal log-likelihood, we evaluate the standard DDPM denoising loss used during training as a single Monte Carlo sample:
\begin{equation}
\mathcal{S}_{\text{ours}} \;=\; -\,\bigl\lVert\, \hat{\epsilon}_\theta(\,x_t,\, t,\, V_t,\, \widehat{V}_{t+\Delta}\,) \;-\; \epsilon \,\bigr\rVert_2^2,
\label{eq:gradcam_ours_scalar}
\end{equation}
with $x_t = \sqrt{\bar\alpha_t}\, T^\star + \sqrt{1-\bar\alpha_t}\, \epsilon$, $t = T/2 = 25$ (the mid-noise level of the training/inference horizon $T=50$), $\epsilon \sim \mathcal{N}(0, I_{12})$ with a deterministic seed, and $T^\star \in \mathbb{R}^{12}$ the GT 6-step ego trajectory normalized by trajectory scale $s=20$.
The conditioning of $\hat{\epsilon}_\theta$ consists of (a) current visual state tokens $V_t$ (DINOv3 cache, $540 \times 896$) and (b) the next visual state latent $\widehat{V}_{t+\Delta}$ predicted by the LLM (1\,s/2\,s/3\,s multi-horizon).
$\mathcal{S}_{\text{ours}}$ is identical to $\lambda_{\text{traj}}\mathcal{L}_{\text{traj}}$ used during training, and the gradient of a single $(t, \epsilon)$ sample is an unbiased estimator of the full training loss gradient.

\noindent\textbf{Token-output baselines (OpenDriveVLA, Impromptu-VLA).}
For models in which the LLM emits the trajectory as a text token sequence, we use the token-level cross-entropy loss with respect to the GT trajectory tokens used during training as is: $\mathcal{S}_{\text{tok}} = -\mathrm{CE}(\text{LLM logits},\, \text{GT trajectory tokens})$.
This matches the native training supervision of those models.

\noindent\textbf{Diffusion-only baseline (Alpamayo-R1).}
Alpamayo-R1 architecturally lacks token-level GT trajectory supervision and uses only a diffusion action head.
In this case, among differentiable scalars over LLM hidden states, we use the logit sum at the \texttt{\textless{}traj\_future\_start\textgreater{}} trigger position closest to the start of trajectory generation as a proxy.
This is not a strict log-likelihood, and measures how much visual context affects the LLM's representation right before trajectory generation.

\subsection{Hook Layer Selection}
\label{sec:gradcam_layer}

For all models, we hook at the same location: the LLM 1st decoder layer pre-hook, i.e., the point where the visual representation has just entered the LLM after the vision encoder (and the projector, when present). This unification ensures that all GradCAM measurements compare gradients at a structurally equivalent stage of the network.

\section{Architecture-Agnostic Evaluation of Visual Grounding}
\label{sec:architecture_agnostic_grounding}

\paragraph{Why a shared behavioral evaluation is needed.}
The GradCAM comparisons use different differentiation targets: a diffusion denoising objective for our model, token cross-entropy for autoregressive baselines, and a proxy logit for Alpamayo-R1 (\cref{sec:gradcam_backprop_target}).
GradCAM localizes sensitivity within each architecture.
Our primary cross-architecture evaluation uses shared pixel interventions and measures their effects directly on predicted trajectories.
Hazard, placebo, and null controls distinguish obstacle-relevant responses from irrelevant visual sensitivity and decoding variability, giving a common behavioral test across architectures.

\paragraph{Common intervention protocol.}
We extend the obstacle-stitching evaluation with placebo and null controls and apply the same protocol to all four models on the full nuScenes validation set ($n=6{,}019$).
Each model receives identical synthesized image pixels for each intervention, while ego status, command, history, and other non-visual context are held fixed.
The hazard arm inserts a vehicle ahead of the ego vehicle; the placebo arm places the same sprite in the sky.
The null arm re-runs the identical, unmodified input to measure the model's own decoding variability.
We compare the resulting trajectories with their unmodified-input counterparts.
The evaluation uses only input interventions and output trajectories, and therefore does not require access to a particular architecture, attention map, or latent representation.

\paragraph{Threshold-free hazard--placebo separation.}
Response rates count the fraction of scenes whose predicted trajectory shortens under each intervention.
Their interpretation depends on the shortening threshold, and model rankings can change when response magnitudes differ.
We therefore report hazard-versus-placebo AUC as the primary comparison, alongside hazard, placebo, and null response rates.
Let $s_i^{\mathrm{haz}}$ and $s_j^{\mathrm{sky}}$ denote signed trajectory-shortening scores relative to the corresponding unmodified trajectories, with larger scores indicating stronger shortening.
The empirical AUC is
\begin{equation}
\operatorname{AUC}
=\frac{1}{n^2}\sum_{i=1}^{n}\sum_{j=1}^{n}
\left[\mathbf{1}\!\left(s_i^{\mathrm{haz}}>s_j^{\mathrm{sky}}\right)
+\frac{1}{2}\mathbf{1}\!\left(s_i^{\mathrm{haz}}=s_j^{\mathrm{sky}}\right)\right].
\label{eq:grounding_auc}
\end{equation}
Thus, AUC estimates the probability that a randomly selected hazard response exceeds a randomly selected placebo response, assigning half credit to ties.
An AUC of $0.5$ indicates chance-level ranking, while $1.0$ indicates perfect separation in the intended direction.
This comparison avoids choosing a single response threshold, although it remains specific to the chosen intervention and response score.

\begin{table}[htbp]
\centering
\small
\caption{Architecture-agnostic grounding evaluation on all $6{,}019$ nuScenes validation scenes, using identical synthesized pixels across models. AUC compares hazard and sky-placebo shortening scores; response-rate columns use the same intervention protocol.}
\label{tab:architecture_agnostic_grounding}
\begin{adjustbox}{max width=\linewidth}
\begin{tabular}{lcccc}
\toprule
Model & Hazard response $\uparrow$ & Sky placebo $\downarrow$ & Same-pixel null $\downarrow$ & AUC $\uparrow$ \\
\midrule
Ours & 63.8\% & 48.3\% & 0.00\% & \textbf{0.628} \\
OpenDriveVLA & 44.9\% & 35.7\% & 0.45\% & 0.542 \\
Impromptu-VLA & 35.7\% & 36.4\% & 0.00\% & 0.478 \\
Alpamayo-R1 & 37.3\% & 49.3\% & 0.00\% & 0.395 \\
\bottomrule
\end{tabular}
\end{adjustbox}
\end{table}

\paragraph{Results and interpretation.}
Our model has the highest hazard--placebo AUC ($0.628$), exceeding OpenDriveVLA ($0.542$), Impromptu-VLA ($0.478$), and Alpamayo-R1 ($0.395$).
The null response is $0.00\%$ for our model, Impromptu-VLA, and Alpamayo-R1, and $0.45\%$ for OpenDriveVLA, indicating that the observed response rates are not explained by these measured null rates alone.
Our model achieves the strongest hazard--placebo separation across the compared architectures, with an AUC advantage of $0.086$ over OpenDriveVLA.
The sky-placebo response rate of $48.3\%$ identifies remaining sensitivity to irrelevant insertions (\cref{sec:limitations}).
Together with trajectory accuracy and safety metrics, the shared-intervention results establish improved discrimination of obstacle relevance in the model's planning response.

\section{Question and Answering Performance.}
\begin{table*}[t]
\centering
\setlength{\tabcolsep}{3pt}
\begin{minipage}[t]{0.48\textwidth}
\centering
\caption{Performance on nu-Caption. \textbf{Bold}=best, \underline{Underline}=second.}
\vspace{4pt}
\resizebox{\linewidth}{!}{%
\begin{tabular}{lccccc}
\toprule
\textbf{Method} & \textbf{BL-1} & \textbf{BL-2} & \textbf{BL-3} & \textbf{BL-4} & \textbf{BERT-S} \\
\midrule
Mini-GPT4         & 15.0 & 6.8  & 3.7  & 2.6  & 84.4 \\
Instruct-BLIP     & 18.7 & 13.4 & 7.4  & 5.2  & 85.9 \\
LLaVA1.5          & 20.0 & 12.1 & 8.6  & 5.4  & 85.0 \\
LLaMA-AdapV2      & 30.2 & 17.3 & 10.4 & 7.5  & 86.5 \\
LiDAR-LLM         & 41.0 & 30.0 & 23.4 & 19.3 & 91.3 \\
OpenDriveVLA-0.5B & \underline{47.2} & \underline{35.8} & \underline{29.4} & \underline{25.2} & \underline{91.9} \\
\midrule
Ours-0.5B         & \textbf{51.2} & \textbf{39.0} & \textbf{31.8} & \textbf{26.9} & \textbf{97.6} \\
\bottomrule
\end{tabular}
\label{tab:main_qa_nucaption}
}
\end{minipage}
\hfill
\begin{minipage}[t]{0.48\textwidth}
\centering
\caption{Performance on Nu-X. \textbf{Bold}=best, \underline{Underline}=second.}
\vspace{4pt}
\resizebox{\linewidth}{!}{%
\begin{tabular}{lcccc}
\toprule
\textbf{Method} & \textbf{CIDEr} & \textbf{BLEU-4} & \textbf{METEOR} & \textbf{ROUGE-L} \\
\midrule
Hint-UniAD        & 21.7 & 4.2 & 12.7 & 27.0 \\
Hint-VAD          & 22.4 & 4.2 & \underline{13.2} & \underline{27.6} \\
GPT-4o            & 19.0 & 4.0 & 10.3 & 24.9 \\
Gemini 1.5        & 17.6 & 3.4 & 9.3  & 23.4 \\
Vote2Cap-DETR     & 15.3 & 2.6 & 10.9 & 24.2 \\
TOD$^3$Cap        & 14.5 & 2.5 & 10.5 & 23.5 \\
OpenDriveVLA-0.5B & \underline{32.3} & \underline{5.4} & 12.5 & \textbf{27.9} \\
\midrule
Ours-0.5B         & \textbf{35.2} & \textbf{5.8} & \textbf{20.2} & 26.2 \\
\bottomrule
\end{tabular}
\label{tab:main_qa_nux}
}
\end{minipage}
\end{table*}

\subsection{Experiments on nu-Caption.}
\label{sec:nucaption_appendix}
As shown in \cref{tab:main_qa_nucaption}, our method also consistently outperforms the baselines, including OpenDriveVLA, on all metrics in driving scene captioning.
Specifically, BLEU-1 through BLEU-4 are recorded as $51.2$, $39.0$, $31.8$, and $26.9$, respectively, showing consistent improvements of $+4.0$, $+3.2$, $+2.4$, and $+1.7$ over OpenDriveVLA \cite{zhou2025opendrivevla}.
In particular, the BERT-Score, which measures semantic similarity, reaches $97.6$, a large gap of $+5.7$ over OpenDriveVLA ($91.9$), suggesting that beyond mere n-gram matching, the LLM understands and describes the driving scene more accurately at the semantic level.
This result supports that our training strategy, which forces attention onto visual tokens through next visual state prediction, recovers not only trajectory planning but also scene understanding ability itself.

\subsection{Experiments on nu-X.}
\label{sec:nux_appendix}
As shown in \cref{tab:main_qa_nux}, our method also shows improvements over the baseline on three out of four metrics on the nu-X benchmark, which requires reasoning about driving scenes.
On CIDEr, we achieve $35.2$, an improvement of $+2.9$ over OpenDriveVLA \cite{zhou2025opendrivevla}, and on BLEU-4 we record $5.8$, an improvement of $+0.4$.
In particular, on METEOR we achieve $20.2$, a large gap of $+7.0$ over Hint-VAD \cite{ding2024hint} ($13.2$), which means that our model produces answers that are semantically richer and more aligned with the reference descriptions.
On ROUGE-L, we record $26.2$, slightly below OpenDriveVLA ($27.9$), but considering the large advantage on CIDEr$\cdot$METEOR, which together comprehensively reflect overall caption quality, we can confirm that our model performs more meaningful reasoning about driving scenes.

\begin{table}[t]
\centering
\caption{Performance on nuScenes-QA. \textbf{Bold}=best, \underline{Underline}=second.}
\resizebox{0.8\columnwidth}{!}{%
\begin{tabular}{lcccccccc}
\toprule
\textbf{Method} & \textbf{Ext} & \textbf{Cnt} & \textbf{Obj} & \textbf{Sts} & \textbf{Cmp} & \textbf{H0} & \textbf{H1} & \textbf{Acc} \\
\midrule
LLaMA-AdapV2      & 19.3 & 2.7  & 7.6  & 10.8 & 1.6  & 15.1 & 4.8  & 9.6  \\
LLaVA1.5          & 45.8 & 7.7  & 7.8  & 9.0  & 52.1 & 25.7 & 41.5 & 26.2 \\
LiDAR-LLM         & 74.5 & 15.0 & 37.8 & 45.9 & 57.8 & -    & -    & 48.6 \\
BEVDet+BUTD       & \underline{83.7} & \underline{20.9} & \underline{48.8} & 52.0 & \underline{67.7} & - & - & \underline{57.0} \\
OpenDriveVLA-0.5B & \textbf{83.9} & \textbf{22.0} & \textbf{50.2} & \textbf{57.0} & \textbf{68.4} & \textbf{62.3} & \textbf{56.5} & \textbf{58.4} \\
\midrule
Ours-0.5B         & 83.7 & 19.5 & 47.1 & \underline{52.8} & 67.3 & \underline{59.9} & \underline{54.7} & 56.4 \\
\bottomrule
\end{tabular}
\label{tab:main_qa_nuscenesqa}
}
\end{table}
\subsection{Experiments on nuScenes-QA.}
\label{sec:main_experiments_nuqa}
As shown in \cref{tab:main_qa_nuscenesqa}, on nuScenes-QA our method is slightly below OpenDriveVLA \cite{zhou2025opendrivevla} in overall accuracy, with a gap of only $2.0$ points ($58.4$ vs.\ $56.4$), and remains within $0.6$ points of BEVDet+BUTD \cite{qian2024nuscenes} ($57.0$).
In detailed categories, we also show nearly the same performance as OpenDriveVLA on \textit{Existence} ($83.7$ vs.\ $83.9$) and \textit{Comparison} ($67.3$ vs.\ $68.4$), and show relatively larger gaps in categories that require fine-grained object identification, such as \textit{Counting} ($-2.5$), \textit{Object} ($-3.1$), and \textit{Status} ($-4.2$).

This performance gap can be explained by the difference in visual representation between our method and OpenDriveVLA.
Since OpenDriveVLA leverages a UniAD BEV encoder \cite{hu2023planning} pretrained on nuScenes, it operates on top of a top-down representation in which the spatial layout of object instances is explicitly encoded.
This is naturally favorable for the Counting/Object/Status categories of nuScenes-QA, which require accurate instance-level identification and spatial relations, such as ``\emph{how many vehicles are visible?}'' or ``\emph{what is the status of the object on the right of the ego?}''.
In contrast, in order to generalize to environments without a publicly available BEV encoder, such as NAVSIM, our method intentionally adopts a perspective-view feature (DINOv3), and the next state prediction objective is also designed to learn planning-relevant scene dynamics rather than fine-grained object identification.
That is, our method pays a slight cost in instance-level perception relative to BEV-based models, but in exchange obtains consistent and large improvements on the essential tasks of a driving VLA, planning and high-level scene understanding.
Moreover, despite the overall nuScenes-QA gap ($-2.0$ points), our accuracy stays within $0.6$ points of BEVDet+BUTD, so our method does not significantly sacrifice perception ability; rather, it shows only a small gap on the instance-level accuracy that BEV pretraining provides.

\section{Ablation Study}
\label{sec:ablation_study}

\subsection{Ablation of the Core Components}
\label{sec:ablation_core_components}
To verify the strengths of our methodology, we removed each of the three core components of our model one by one and analyzed their contributions (\cref{tab:ablation_study}).

\noindent\textbf{Effect of Next State Prediction.}
First, we removed \emph{next state prediction}, which trains the LLM to predict the future ($\Delta = 3$\,s) visual state, and replaced it with \emph{current state reconstruction} as the training objective.
As a result, ADE degraded by $+0.01$, from $0.06 \rightarrow 0.07$.
This comparison measures the contribution of future-state supervision while retaining the visual-state IK interface.

\noindent\textbf{Effect of the IK Network.}
Second, we removed the \emph{IK Network} that takes the current/next visual state as input and decodes the trajectory, and replaced it with a small MLP head applied directly to the LLM's output hidden state.
As a result, ADE degraded by $+0.25$, from $0.06 \rightarrow 0.31$, i.e., a nearly \textbf{$5\times$} larger gap.
This suggests that when the trajectory is regressed directly from the LLM hidden state, the model is again exposed to the shortcut learning problem.
In contrast, our IK Network separates trajectory generation into a separate module and restricts its inputs \emph{only to the visual state}, thereby forcing the LLM to learn scene understanding before memorizing trajectories.

\noindent\textbf{Architecture of the IK Network.}
Third, to verify the structural design choices of the IK Network itself, we replaced the diffusion-based IK Network with a \emph{non-generative} variant.
Specifically, we changed it to a structure that takes the current/next visual state as keys and values, performs cross-attention pooling with a learnable query, and then \emph{directly} regresses the trajectory through an MLP regression head.
As a result, ADE degraded by $+0.10$, from $0.06 \rightarrow 0.16$.
This suggests two things.
First, since this variant still preserves the core design of next state prediction and a visual-only IK module, an ADE of $0.16$ is still overwhelmingly better than \emph{w/o IK Network} ($0.31$).
The advantage persists with deterministic decoding, showing that the benefit of the state-conditioned IK design extends beyond the diffusion decoder.
Second, even within the same structure, implementing the IK module as a \emph{generative diffusion} rather than a deterministic regression elevates trajectory precision one step further, which is a natural result given that the trajectory distribution is inherently multi-modal and stochastic.
\cref{tab:ablation_study} summarizes these results.

\begin{table}[t]
\caption{
\textbf{Ablation study on nuScenes \emph{val}.}
\textbf{Bold}=best.
}
\centering
\begin{adjustbox}{max width=\linewidth}
\begin{tabular}{lc}
\toprule
\textbf{Method} & \textbf{ADE (m)}$\downarrow$  \\
\midrule
\textit{w/o Next State Prediction} (current state pred.)         & 0.07 \\
\textit{w/o IK Network} (direct MLP traj.\ pred.)                & 0.31 \\
\textit{w/o Diffusion in IK} (cross-attn pooling + MLP)          & 0.16 \\
\midrule
Ours (full model)                                                & \textbf{0.06} \\
\bottomrule
\end{tabular}
\end{adjustbox}
\label{tab:ablation_study}
\end{table}

\subsection{Disentangling Next-State Prediction from the Visual Diffusion Decoder}
\label{sec:nsp_ik_disentanglement}

\paragraph{Motivation and controlled protocol.}
The component ablation (\cref{tab:ablation_study}) shows a small ADE increase when NSP is replaced by current-state reconstruction ($0.06\rightarrow0.07$\,m), compared with the larger degradation when the IK Network is removed ($0.06\rightarrow0.31$\,m).
To isolate NSP's contribution to visual grounding, we hold the diffusion decoder fixed and measure both trajectory accuracy and hazard-specific response.
We therefore conduct a separate, unified four-epoch ablation on nuScenes, training every variant from scratch with its ablation active from the beginning and evaluating on all $6{,}019$ validation samples.
The main-table training comprises four nuScenes epochs plus one additional epoch on the language dataset; this ablation omits the additional language epoch to remove its contribution from the comparison.
All variants use the same diffusion IK head and training schedule; only the conditioning configuration and corresponding state-supervision target change.
Because this series excludes the additional language training used for the main-table models, its absolute results are reported separately; comparisons between conditioning variants use only this unified series.

\paragraph{Conditioning variants.}
Row (0) zeros the cross-attention keys and values and provides neither visual conditioning nor ego status or command.
Row (0+) retains this absence of visual conditioning but injects ego status and command directly into the decoder.
Row (i) uses current-state reconstruction instead of NSP, providing current-state information only; the decoder architecture and conditioning-token count are matched to row (ii).
Row (ii) is our predicted-future configuration, conditioned on $(V_t,\hat V_{t+\Delta})$.
Row (iii) replaces the predicted future with the ground-truth future visual feature, using $(V_t,V_{t+\Delta}^{\mathrm{GT}})$ during both training and evaluation.
Row (iii) requires an unavailable future observation at deployment and is included as a diagnostic.

\paragraph{Hazard and placebo measurements.}
For each scene, we compare an unmodified input with an input containing a synthetic vehicle ahead, keeping the inference seed fixed between the paired runs, following the obstacle-stitching protocol in \cref{tab:stitching_exp}.
The hazard deceleration rate is the fraction of scenes in which the predicted trajectory shortens after insertion.
We repeat the comparison with the same sprite inserted into the sky as a placebo.
The hazard--placebo gap measures response selectivity in this intervention: a large hazard response accompanied by a similarly large sky response is weak evidence of hazard-specific grounding.
Mean $\Delta$displacement is reported for the hazard intervention, with negative values indicating shorter trajectories.

\begin{table}[htbp]
\centering
\small
\caption{Controlled conditioning ablation on nuScenes validation ($n=6{,}019$), with each variant trained from scratch for four epochs. L2 averages are reported under the ST-P3 and UniAD evaluation protocols. Hazard and sky rates are percentages; their difference is in percentage points. Lower L2 and a larger hazard--sky gap are desirable; a larger displacement reduction is not necessarily better.}
\label{tab:nsp_ik_conditioning}
\begin{adjustbox}{max width=\linewidth}
\begin{tabular}{lccccc}
\toprule
IK conditioning & L2 avg. (m) $\downarrow$ & Hazard (\%) & Sky (\%) & Gap (pp) $\uparrow$ & Mean $\Delta$displ. (m) \\
 & ST-P3 / UniAD & & & & \\
\midrule
(0) No conditioning (K/V zeroed) & 5.58 / 8.62 & 0.0 & 0.0 & 0.0 & 0.000 \\
(0+) No vision; direct ego/command & 0.60 / 1.11 & 0.0 & 0.0 & 0.0 & 0.000 \\
(i) Current-state information only & 0.55 / 1.01 & 71.3 & 65.5 & 5.8 & $-0.21$ \\
(ii) $V_t+\hat V_{t+\Delta}$ (ours) & \textbf{0.53 / 0.98} & 82.8 & 45.2 & 37.6 & $-0.62$ \\
(iii) $V_t+V_{t+\Delta}^{\mathrm{GT}}$ & 2.58 / 4.05 & 95.4 & 46.1 & 49.3 & $-8.71$ \\
\bottomrule
\end{tabular}
\end{adjustbox}
\end{table}

\paragraph{Visual conditioning and accuracy without grounding.}
Without any conditioning, row (0) has L2 errors of $5.58/8.62$\,m.
Providing only ego status and command in row (0+) reduces these errors to $0.60/1.11$\,m despite zero response to either visual intervention.
Thus, relatively low trajectory error is attainable without visual responsiveness, making L2 alone insufficient to establish grounding.
Adding current-state conditioning in row (i) improves accuracy and introduces a visual response, but its small hazard--placebo gap indicates limited selectivity.

\paragraph{NSP adds response selectivity beyond the same diffusion decoder.}
The controlled comparison between rows (i) and (ii) holds the diffusion architecture and token count fixed.
NSP improves L2 from $0.55/1.01$ to $0.53/0.98$\,m, raises the hazard deceleration rate from $71.3\%$ to $82.8\%$, and reduces the sky-placebo rate from $65.5\%$ to $45.2\%$.
The hazard--placebo gap increases from $5.8$ to $37.6$ percentage points, approximately $6.5\times$, while the magnitude of mean hazard-induced shortening increases from $0.21$ to $0.62$\,m.
With the diffusion decoder and token count fixed, this $6.5\times$ increase establishes NSP's contribution to hazard-specific response selectivity.
The direction of the accuracy change agrees with the component ablation's ADE increase from $0.06$ to $0.07$\,m when NSP is removed (\cref{tab:ablation_study}).
Among the deployable configurations, row (ii) combines the lowest L2 error with the largest hazard--placebo gap.
The improvement combines a higher hazard response with a lower placebo response; the remaining $45.2\%$ placebo rate quantifies residual sensitivity to irrelevant visual changes.

\paragraph{Predicted future states preserve planning-relevant context.}
Row (iii) achieves a high hazard deceleration rate ($95.4\%$), but its L2 errors rise to $2.58/4.05$\,m and the mean displacement change reaches $-8.71$\,m.
This combination is consistent with excessive trajectory shortening rather than improved planning, despite a large hazard--placebo gap.
Thus, raw future visual features do not serve as an upper bound for this decoder, and intervention response rates must be interpreted alongside trajectory accuracy and response magnitude.

To examine the difference between predicted and ground-truth future representations, we report linear probes for ego speed and command in \cref{tab:nsp_future_probes}.
The predicted representation supports ego-speed prediction with $R^2=0.99$ and command classification with $98.5\%$ accuracy, compared with $0.33$ and $79.5\%$ for the ground-truth future feature.
The latter command accuracy is below the $88.2\%$ majority-class baseline.
These probes show that ego-kinematic and goal information is substantially more linearly accessible in the predicted state than in the ground-truth visual feature.
The NSP predictor receives $(E,C)$, while the IK decoder has no direct access to them; replacing its predicted state can therefore change the accessibility of task information even when the replacement is an actual future observation.
The probes measure the linear accessibility of task information, providing a representation-level explanation consistent with the trajectory degradation in row (iii).
Our aim is to suppress planning that bypasses visual evidence, while retaining the ego and command information needed for the requested maneuver.

\begin{table}[htbp]
\centering
\small
\caption{Linear probing of predicted and ground-truth future features. The input-independent reference is a mean predictor for ego speed and a majority-class predictor for command.}
\label{tab:nsp_future_probes}
\begin{tabular}{lccc}
\toprule
Probe target & Predicted $\hat V_{t+\Delta}$ & GT $V_{t+\Delta}$ & Input-independent baseline \\
\midrule
Ego speed ($R^2$) & 0.99 & 0.33 & 0 \\
Command accuracy & 98.5\% & 79.5\% & 88.2\% \\
\bottomrule
\end{tabular}
\end{table}

\subsection{Contribution of QA Co-Training to Trajectory Planning}
\label{sec:qa_planning_ablation}

\paragraph{Iteration-matched ablation.}
To evaluate whether the question-answering component benefits planning, we remove QA multi-task supervision entirely and retrain with the number of optimization iterations matched to the full model ($140{,}000$ steps).
Both models are evaluated on the full nuScenes validation set ($n=6{,}019$) using the ST-P3 planning metrics.
This comparison controls the number of optimization steps, so the QA model's advantage cannot be explained simply by a larger iteration budget.
The controlled quantity is the optimization-step budget; QA supplies the additional supervision being evaluated.

\begin{table}[htbp]
\centering
\small
\caption{Effect of QA co-training on nuScenes validation ($n=6{,}019$). Both models are trained for $140{,}000$ optimization steps. Lower is better for both ST-P3 metrics.}
\label{tab:qa_planning_ablation}
\begin{tabular}{lcc}
\toprule
Model & L2 avg. (m) $\downarrow$ & Collision avg. (\%) $\downarrow$ \\
\midrule
Ours (with QA co-training) & \textbf{0.06} & \textbf{0.09} \\
QA removed (iterations matched) & 0.38 & 0.18 \\
\bottomrule
\end{tabular}
\end{table}

Removing QA increases average L2 error by $0.32$\,m, from $0.06$ to $0.38$\,m, and doubles the reported collision rate from $0.09\%$ to $0.18\%$.
QA co-training therefore improves both trajectory accuracy and collision rate under the matched iteration budget.

\paragraph{Training-time transfer through the predicted state.}
QA provides auxiliary supervision to the shared LLM during training.
At planning inference, the IK Network consumes only the current and predicted future visual states, $(V_t,\hat V_{t+\Delta})$; generated QA answers are not inputs to the trajectory decoder.
QA updates can therefore improve the representation used to predict $\hat V_{t+\Delta}$, through which the shared LLM influences the trajectory.
QA transfers to planning through the shared representation learned during training, with no generated answer in the inference-time trajectory path.
The route command remains an upstream input to future-state prediction (\cref{sec:methods,sec:appendix_arch}).

The iteration-matched ablation establishes QA co-training as a substantial contributor to planning performance, consistent with its role in learning the shared future-state representation.
The obstacle-stitching evaluation and position--size controls provide complementary evidence about visual use (\cref{sec:stitching,sec:stitching_spatial}).
These results also make QA co-training part of the explanation for the full model's planning performance, alongside NSP and the IK decoder.

\section{Inference Speed}
\label{sec:inference_speed}

\noindent\textbf{Setup.}
We measure end-to-end inference latency of our $0.5$B model on a single NVIDIA RTX A6000 over the nuScenes \emph{val set} with three front cameras.
Following common practice in VLA latency reporting, we decompose the forward pass into (i) the visual encoder (DINOv3 ViT-S/16 over $3$ cameras at $288{\times}160$) and (ii) the LLM together with the IK Network ($50$-step DDPM sampling).
Each stage is timed independently with \texttt{torch.cuda.synchronize()} after $10$\,--\,$20$ warm-up samples.

\noindent\textbf{Stage breakdown and end-to-end latency.}
\cref{tab:inference_speed} summarises the per-stage latency.
The vision encoder takes $20.7$\,ms on average and the LLM\,+\,IK stage takes $323.7$\,ms, giving an end-to-end latency of $\sim 344$\,ms ($2.9$\,FPS) when the camera stream is fed directly to GPU memory (the $\sim 515$\,ms JPEG disk-load cost in our offline benchmark is an artefact of file I/O and does not appear in a real ISP\,$\rightarrow$\,GPU pipeline).

\begin{table}[htbp]
\centering
\caption{
Inference speed of our 0.5B model on a single NVIDIA RTX A6000, measured on the nuScenes validation set with three front cameras and DDPM $50$-step IK sampling.
End-to-end is the GPU-only sum (vision encoder + LLM + IK).
For context, we list reported latency of representative Driving VLAs and large VLMs.
}
\label{tab:inference_speed}
\setlength{\tabcolsep}{6pt}
\resizebox{\linewidth}{!}{%
\begin{tabular}{llcc}
\toprule
\textbf{Model / stage} & \textbf{Hardware} & \textbf{Latency} & \textbf{Throughput} \\
\midrule
\multicolumn{4}{l}{\emph{Per-stage breakdown of our model}} \\
~~Vision encoder (DINOv3 ViT-S/16, 3\,cam)         & A6000 (fp16) & $20.7$\,ms  & --             \\
~~LLM (Qwen2.5-0.5B) + IK diffusion ($50$ steps)   & A6000 (fp16) & $323.7$\,ms & --             \\
~~\textbf{End-to-end (vision + LLM + IK)}          & A6000 (fp16) & $\sim 344$\,ms & $2.9$\,FPS \\
\midrule
\multicolumn{4}{l}{\emph{Reported latency of other Driving VLAs and VLMs}} \\
VLA-MP \cite{ge2025vlamp}                  & RTX 3090     & $125$\,ms   & $8.0$\,FPS \\
OpenDriveVLA-0.5B \cite{zhou2025opendrivevla}     & A100 (bf16)  & $1.36$\,s   & $0.74$\,FPS \\
OpenDriveVLA-3B   \cite{zhou2025opendrivevla}     & A100 (bf16)  & $1.85$\,s   & $0.54$\,FPS \\
OpenDriveVLA-7B   \cite{zhou2025opendrivevla}     & A100 (bf16)  & $1.74$\,s   & $0.57$\,FPS \\
DriveVLM \cite{tian2024drivevlm}                  & --           & $\sim 1.5$\,s & $0.67$\,FPS \\
AutoVLA \cite{zhou2025autovla}                    & --           & $\sim 1.0$\,s & $\sim 1$\,FPS \\
Qwen-2.5-VL-7B \cite{bai2025qwen25vl}            & H100         & $8.8$\,s    & $0.11$\,FPS \\
LLaMA-3.2-11B  \cite{grattafiori2024llama}       & H100         & $7.5$\,s    & $0.13$\,FPS \\
GPT-4o (closed)                                   & API          & $12.1$\,s   & $0.083$\,FPS \\
\bottomrule
\end{tabular}}
\end{table}

\noindent\textbf{Comparison to other VLAs.}
It is worth noting that our $\sim 344$\,ms is achieved on an RTX A6000, a workstation-class GPU that is markedly slower than the data-center A100/H100s used by the baselines below; despite this hardware disadvantage, our model has competitive latency.
At the same $0.5$B backbone, OpenDriveVLA reports $1.36$\,s on A100 (bf16), so our model is roughly $4\times$ faster while running on the slower GPU.
At the VLA category level, mainstream Driving VLAs (DriveVLM, OpenDriveVLA-3B/7B, AutoVLA) sit around $1$\,--\,$2$\,s on A100, and large general-purpose VLMs (Qwen-2.5-VL-7B, LLaMA-3.2-11B/90B, GPT-4o, GPT-5) range from several seconds to tens of seconds even on H100 or larger clusters.
Sub-$500$\,ms reports in the VLA literature are rare; ours is one of the few in this range, alongside specialised systems such as VLA-MP ($125$\,ms on RTX 3090).
This suggests that, on top of the accuracy gains in the main paper, the proposed architecture is also a practical choice for latency-sensitive deployment.
The dominant cost is the $50$-step DDPM sampling inside the IK Network, leaving room for further reduction via fewer diffusion steps, distillation, or visual-token pruning.

\section{Limitations and Future Work.}
\label{sec:limitations}
We have not compared the LLM future-state generator with a matched non-language alternative, or jointly removed ego status and command in a factorial experiment.
These comparisons would further characterize the role of the LLM backbone and the interaction between ego status and command in future-state prediction (\cref{sec:methods}).

Our method shows somewhat lower performance than BEV-based VLAs on a few categories of nuScenes-QA that require instance-level perception (Counting, Object, etc.),
which is the result of a trade-off in which we intentionally chose a perspective-view representation in order to generalize to environments where no pretrained BEV encoder is provided (e.g., NAVSIM).

In addition, while our paired stitching analysis (\cref{sec:stitching_spatial}) shows that the recovered visual grounding is genuinely position-aware (Far vs.\ SkyFar discrimination at $65.5\%$, paired mean $-0.64$\,m), the precision of this grounding still has room to grow---the magnitude of the response also scales with sprite size, and a fraction of off-road placements still elicit non-trivial reactions.
A natural direction for sharpening this signal is supervision that explicitly disentangles the position, size, and meaning of obstacles (e.g., using the ego-frame coordinates of synthesized obstacles as auxiliary outputs), and additional research in this direction is needed.

Moreover, our experiments are limited to open-loop and short-term closed-loop benchmarks.
Validation on long-term closed-loop environments such as Bench2Drive \cite{jia2024bench2drive},
and extending the inverse-kinematics redefinition of this work to other end-to-end embodied policies beyond autonomous driving, would be interesting follow-up directions.

\section*{NeurIPS Paper Checklist}

\begin{enumerate}

\item {\bf Claims}
    \item[] Question: Do the main claims made in the abstract and introduction accurately reflect the paper's contributions and scope?
    \item[] Answer: \answerYes{} 
    \item[] Justification: The abstract and Section~\ref{sec:intro} state our two main claims---(i) that existing Driving VLAs ignore visual tokens and learn shortcuts, and (ii) that our next visual state prediction with an Inverse Kinematics Network mitigates this and improves trajectory planning. Both claims are supported by experiments in Section~\ref{sec:experiments} (e.g., GradCAM grounding analysis, NAVSIM SOTA, and ablations).

\item {\bf Limitations}
    \item[] Question: Does the paper discuss the limitations of the work performed by the authors?
    \item[] Answer: \answerYes{} 
    \item[] Justification: We discuss limitations in Section~\ref{sec:main_experiments_nuqa} (slight under-performance on instance-level perception tasks such as nuScenes-QA Counting/Object compared to BEV-based models) and in Appendix. We also note that experiments are limited to nuScenes and NAVSIM benchmarks, and that the Inverse Kinematics Network requires a forward-pass diffusion sampling step that adds inference overhead.

\item {\bf Theory assumptions and proofs}
    \item[] Question: For each theoretical result, does the paper provide the full set of assumptions and a complete (and correct) proof?
    \item[] Answer: \answerYes{} 
    \item[] Justification: Section~\ref{sec:motivations} states the bias--variance decomposition of the trajectory loss together with its assumptions (the majority-regime conditional independence and zero-mean noise independent of the inputs; \cref{eq:stylized_independence}), and \cref{sec:formulation} gives numbered step-by-step derivations for direct regression, next visual state prediction, and the IK Network. The analysis is a decomposition under these stated, idealized assumptions, whose empirical calibration is reported in \cref{sec:eps_sigma_estimates}.

    \item {\bf Experimental result reproducibility}
    \item[] Question: Does the paper fully disclose all the information needed to reproduce the main experimental results of the paper to the extent that it affects the main claims and/or conclusions of the paper (regardless of whether the code and data are provided or not)?
    \item[] Answer: \answerYes{} 
    \item[] Justification: We describe the full architecture in Section~\ref{sec:methods}, training objectives and loss formulations, dataset splits (nuScenes train/val, NAVSIM splits), and key hyperparameters (e.g., DDPM cosine schedule with 50 sampling steps, IK Network dimensions) in Section~\ref{sec:experiments} and Appendices~\ref{sec:appendix_arch}--\ref{sec:appendix_training}. Code will be publicly available at \url{https://github.com/engineerJPark/DriveIK}.

\item {\bf Open access to data and code}
    \item[] Question: Does the paper provide open access to the data and code, with sufficient instructions to faithfully reproduce the main experimental results, as described in supplemental material?
    \item[] Answer: \answerYes{} 
    \item[] Justification: Training and inference code, with a README describing dataset preparation, training commands, and inference protocols for reproducing the main results, will be publicly available at \url{https://github.com/engineerJPark/DriveIK}. All datasets used (nuScenes, NAVSIM, nuCaption, NuX) are publicly available, and our preprocessing follows the original splits without modification.

\item {\bf Experimental setting/details}
    \item[] Question: Does the paper specify all the training and test details (e.g., data splits, hyperparameters, how they were chosen, type of optimizer) necessary to understand the results?
    \item[] Answer: \answerYes{} 
    \item[] Justification: Section~\ref{sec:experiments} report training details including optimizer (AdamW), learning rates, batch sizes, training schedules (e.g., 10/12/20-epoch stages), and IK Network hyperparameters (cosine $\beta$-schedule, 50 sampling steps). Dataset splits follow the standard nuScenes train/val and NAVSIM v1/v2 protocols.

\item {\bf Experiment statistical significance}
    \item[] Question: Does the paper report error bars suitably and correctly defined or other appropriate information about the statistical significance of the experiments?
    \item[] Answer: \answerYes{} 
    \item[] Justification: Statistical significance is reported for our key analyses: GradCAM--object overlap differences are accompanied by 95\% bootstrap confidence intervals (Appendix~\ref{sec:gradcam_methodology}, Table~\ref{tab:gradcam_overlap}). For trajectory planning metrics on nuScenes/NAVSIM (Tables~\ref{tab:main_exp}, \ref{tab:navsim_v1_exp}, \ref{tab:navsim_v2_exp}), we follow the standard single-run reporting protocol of prior work in this benchmark, as multi-seed evaluation with billion-parameter VLAs is computationally prohibitive.

\item {\bf Experiments compute resources}
    \item[] Question: For each experiment, does the paper provide sufficient information on the computer resources (type of compute workers, memory, time of execution) needed to reproduce the experiments?
    \item[] Answer: \answerYes{} 
    \item[] Justification: Training was conducted on 4 NVIDIA H100 80GB GPUs. The full 0.5B model takes approximately 12 hours per training stage.
    
\item {\bf Code of ethics}
    \item[] Question: Does the research conducted in the paper conform, in every respect, with the NeurIPS Code of Ethics \url{https://neurips.cc/public/EthicsGuidelines}?
    \item[] Answer: \answerYes{} 
    \item[] Justification: This work conforms to the NeurIPS Code of Ethics. We use only publicly available autonomous driving datasets (nuScenes, NAVSIM, nuCaption, NuX) under their respective research licenses, do not collect new human-subjects data, and do not deploy our model in any safety-critical real-world setting.

\item {\bf Broader impacts}
    \item[] Question: Does the paper discuss both potential positive societal impacts and negative societal impacts of the work performed?
    \item[] Answer: \answerYes{} 
    \item[] Justification: We discuss broader impacts in Section~\ref{sec:conclusion}. Positive impacts include improved planning safety through stronger visual grounding in autonomous driving systems. Potential negative impacts include over-reliance on imperfect VLA models in real-world deployment; we emphasize that our results are obtained on offline benchmarks and that real-world deployment requires additional safety validation beyond the scope of this paper.
    
\item {\bf Safeguards}
    \item[] Question: Does the paper describe safeguards that have been put in place for responsible release of data or models that have a high risk for misuse (e.g., pre-trained language models, image generators, or scraped datasets)?
    \item[] Answer: \answerNA{} 
    \item[] Justification: The paper poses no such risks. It does not release pre-trained language models, image generators, or scraped datasets; the released artifact is a trajectory planning model trained on existing public driving benchmarks.

\item {\bf Licenses for existing assets}
    \item[] Question: Are the creators or original owners of assets (e.g., code, data, models), used in the paper, properly credited and are the license and terms of use explicitly mentioned and properly respected?
    \item[] Answer: \answerYes{} 
    \item[] Justification: We cite the original papers and respect the licenses of all datasets and pretrained models used: nuScenes (CC BY-NC-SA 4.0), NAVSIM (Apache 2.0), nuCaption~\cite{yang2025lidar}, NuX~\cite{ding2024hint}, DINOv3~\cite{simeoni2025dinov3}, Qwen2.5 backbones, and the OpenDriveVLA codebase~\cite{zhou2025opendrivevla} which we re-implement for the NAVSIM baseline.

\item {\bf New assets}
    \item[] Question: Are new assets introduced in the paper well documented and is the documentation provided alongside the assets?
    \item[] Answer: \answerYes{} 
    \item[] Justification: Training and inference code and trained model checkpoints will be publicly available at \url{https://github.com/engineerJPark/DriveIK}, with a README describing the directory structure, environment setup, dataset preparation, and reproduction commands for the main results. No new datasets are introduced.

\item {\bf Crowdsourcing and research with human subjects}
    \item[] Question: For crowdsourcing experiments and research with human subjects, does the paper include the full text of instructions given to participants and screenshots, if applicable, as well as details about compensation (if any)? 
    \item[] Answer: \answerNA{} 
    \item[] Justification: This work does not involve crowdsourcing or research with human subjects. All experiments use existing public driving benchmarks.

\item {\bf Institutional review board (IRB) approvals or equivalent for research with human subjects}
    \item[] Question: Does the paper describe potential risks incurred by study participants, whether such risks were disclosed to the subjects, and whether Institutional Review Board (IRB) approvals (or an equivalent approval/review based on the requirements of your country or institution) were obtained?
    \item[] Answer: \answerNA{} 
    \item[] Justification: This work does not involve research with human subjects.

\item {\bf Declaration of LLM usage}
    \item[] Question: Does the paper describe the usage of LLMs if it is an important, original, or non-standard component of the core methods in this research? Note that if the LLM is used only for writing, editing, or formatting purposes and does \emph{not} impact the core methodology, scientific rigor, or originality of the research, declaration is not required.
    \item[] Answer: \answerNA{} 
    \item[] Justification: Our method fine-tunes a standard pretrained LLM (Qwen2.5-0.5B) as a backbone, which is a conventional component in Driving VLA literature rather than an original or non-standard use of LLMs. Any use of LLMs for editorial purposes (e.g., grammar checking) does not affect the core methodology and per the NeurIPS policy does not require declaration.

\end{enumerate}

\end{document}